\documentclass{article}

\usepackage{microtype}
\usepackage{graphicx}
\usepackage{subfigure}
\usepackage{booktabs} % for professional tables

\usepackage{hyperref}

\usepackage[accepted]{icml2024}

\usepackage{amsmath}
\usepackage{amssymb}
\usepackage{mathtools}
\usepackage{amsthm}

\usepackage[capitalize,noabbrev]{cleveref}

\DeclareMathOperator*{\argmax}{arg\,max}
\usepackage{enumitem}
\usepackage{caption}
\usepackage{xcolor}
\usepackage{colortbl}
\usepackage[normalem]{ulem}
\theoremstyle{plain}

\theoremstyle{definition}

\theoremstyle{remark}

\usepackage[textsize=tiny]{todonotes}

\icmltitlerunning{Harmony in Diversity: Merging Neural Networks with Canonical Correlation Analysis}

\begin{document}
\newcommand{\texttildee}{\raisebox{0.5ex}{\texttildelow}}

\twocolumn[
\icmltitle{Harmony in Diversity: Merging Neural Networks with\\ Canonical Correlation Analysis}

% It is OKAY to include author information, even for blind
% submissions: the style file will automatically remove it for you
% unless you've provided the [accepted] option to the icml2024
% package.

% List of affiliations: The first argument should be a (short)
% identifier you will use later to specify author affiliations
% Academic affiliations should list Department, University, City, Region, Country
% Industry affiliations should list Company, City, Region, Country

% You can specify symbols, otherwise they are numbered in order.
% Ideally, you should not use this facility. Affiliations will be numbered
% in order of appearance and this is the preferred way.
\icmlsetsymbol{equal}{*}

\begin{icmlauthorlist}
\icmlauthor{Stefan Horoi}{udem,mila}
\icmlauthor{Albert Manuel Orozco Camacho}{concordia,mila}
\icmlauthor{Eugene Belilovsky}{concordia,mila}
\icmlauthor{Guy Wolf}{udem,mila}
\end{icmlauthorlist}

\icmlaffiliation{udem}{Université de Montréal}
\icmlaffiliation{concordia}{Concordia University}
\icmlaffiliation{mila}{Mila - Quebec AI Institute}

\icmlcorrespondingauthor{Stefan Horoi}{stefan.horoi@umontreal.ca}
% \icmlcorrespondingauthor{Albert Manuel Orozco Camacho}{albert.orozcocamacho@concordia.ca}
\icmlcorrespondingauthor{Guy Wolf}{guy.wolf@umontreal.ca}
\icmlcorrespondingauthor{Eugene Belilovsky}{eugene.belilovsky@concordia.ca}

% \centerline{\texttt{\{stefan.horoi, guy.wolf\}@umontreal.ca}}\\
% \centerline{\texttt{\{albert.orozcocamacho, eugene.belilovsky\}@concordia.ca}}\\

% You may provide any keywords that you
% find helpful for describing your paper; these are used to populate
% the "keywords" metadata in the PDF but will not be shown in the document
\icmlkeywords{Model Merging, Model Fusion, Model Alignment, Linear Mode Connectivity, Canonical Correlation Analysis, Machine Learning, ICML}

\vskip 0.3in]

% this must go after the closing bracket ] following \twocolumn[ ...

% This command actually creates the footnote in the first column
% listing the affiliations and the copyright notice.
% The command takes one argument, which is text to display at the start of the footnote.
% The \icmlEqualContribution command is standard text for equal contribution.
% Remove it (just {}) if you do not need this facility.

\printAffiliationsAndNotice{}  % leave blank if no need to mention equal contribution
% \printAffiliationsAndNotice{\icmlEqualContribution} % otherwise use the standard text.

\begin{abstract}
\vspace{-4pt}
Combining the predictions of multiple trained models through ensembling is generally a good way to improve accuracy by leveraging the different learned features of the models, however it comes with high computational and storage costs. Model fusion, the act of merging multiple models into one by combining their parameters reduces these costs but doesn't work as well in practice. Indeed, neural network loss landscapes are high-dimensional and non-convex and the minima found through learning are typically separated by high loss barriers. Numerous recent works have been focused on finding permutations matching one network features to the features of a second one, lowering the loss barrier on the linear path between them in parameter space. However, permutations are restrictive since they assume a one-to-one mapping between the different models' neurons exists. We propose a new model merging algorithm, CCA Merge, which is based on Canonical Correlation Analysis and aims to maximize the correlations between linear combinations of the model features. We show that our alignment method leads to better performances than past methods when averaging models trained on the same, or differing data splits. We also extend this analysis into the harder setting where more than 2 models are merged, and we find that CCA Merge works significantly better than past methods.
% Our code is publicly available at \url{https://github.com/shoroi/align-n-merge}
\footnote{Our code is publicly available at \url{https://github.com/shoroi/align-n-merge}}
\vspace{-10pt}
\end{abstract}
% \vspace{-10pt}
\section{Introduction}\label{s:intro}
\vspace{-5pt}
\begin{figure}[t]
\begin{center}
\includegraphics[width=0.47\textwidth]{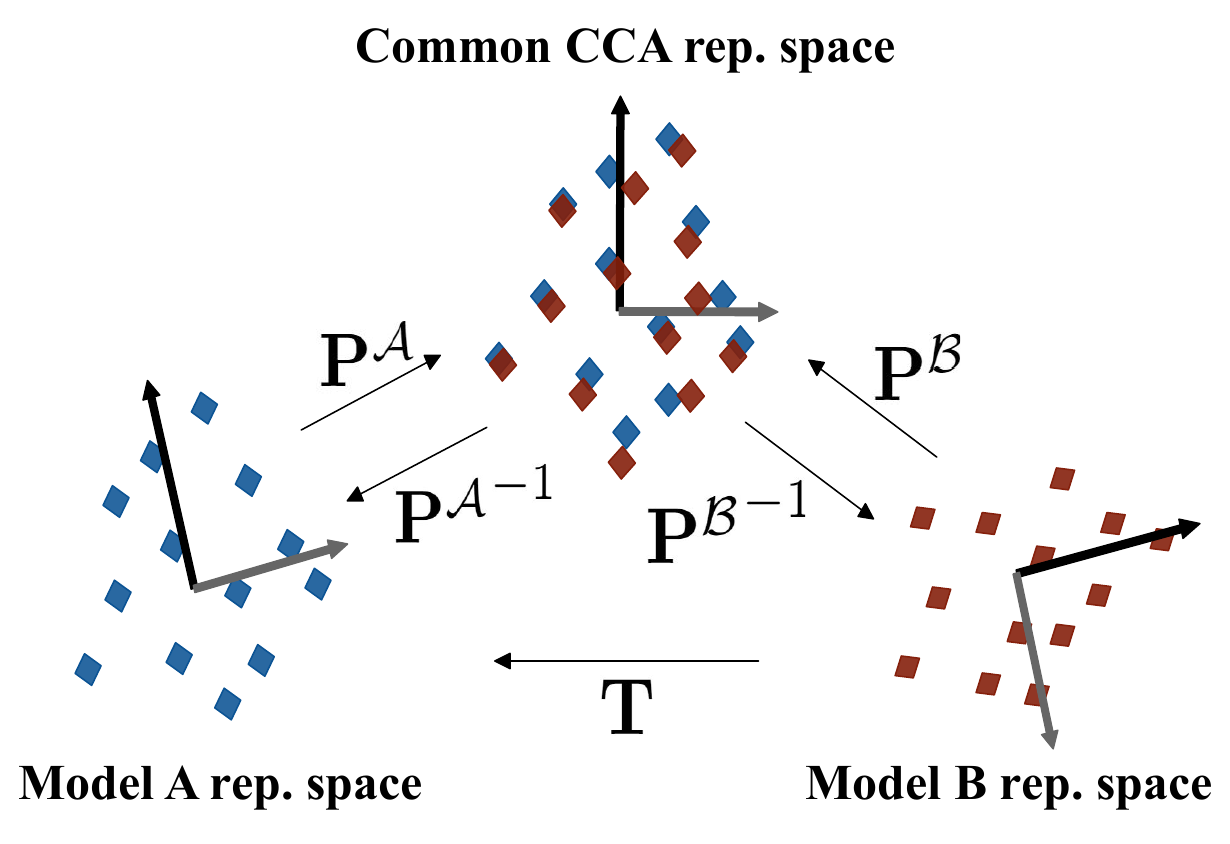}
\end{center}
\vspace{-5pt}
\caption{\textbf{Visual representation of using CCA Merge to align two models.} Canonical Correlation Analysis is used to find a common representation space where orthogonal linear combinations of the features (neurons) from $\mathcal{A}$ and $\mathcal{B}$ are maximally correlated. The linear transformation $\mathbf{P}^\mathcal{A}$ (resp. $\mathbf{P}^\mathcal{B}$) and its inverse can be used to go from the representation space of model $\mathcal{A}$ (resp. model $\mathcal{B}$) to this common representation space and back. By applying $\mathbf{P}^\mathcal{B}$ first and then $\mathbf{P}^{\mathcal{A}^{-1}}$ we can align the representations of model $\mathcal{B}$ to those of model $\mathcal{A}$. Applying the same transformation directly to the parameters of model $\mathcal{B}$ effectively aligns the two models, thus allowing their merging.}
\vspace{-15pt}
\label{fig:cca_merge_fig}
\end{figure}

A classical idea for improving the predictive performance and robustness of machine learning models is to use multiple trained models simultaneously. Each model might learn to extract different, complementary pieces of information from the input data, and combining the models would take advantage of this complementarity; thus benefiting the final performance \citep{ho1995random}.
One of the simplest ways to implement this idea is to use ensembles, where multiple models are trained on a given task and their outputs are combined through averaging or majority vote at inference \citep{ho1995random,lobacheva2020power}. While this method yields good results, it comes with the disadvantages of having to store in memory the parameters of multiple models and having to run each of them individually at inference time, resulting in high storage and computational costs, particularly in the case of neural networks (NNs). 
Another way of leveraging multiple models to improve predictive performance is to combine the different sets of parameters into a single model. This is typically done through averaging or interpolation in the parameter space of the models. After this ``fusion'' only a single model remains which will be used at inference time; therefore, the storage and computational costs are minimized, being the same as for a single model. The downside of such model fusion methods is that existing methods are not robust and typically do not perform as well in practice as ensembling \citep{stoica2024zipit}.  Neural networks are highly over-parameterized for the task they solve \citep{arpit2017_memorization} and their loss landscapes are high-dimensional and non-convex objects which are still somewhat poorly understood despite many recent works shedding light on some of their characteristics \citep{goodfellow2015_qualitatively, keskar2017large_batch, Goldstein_LL_NIPS2018_7875, horoi2022_exploring}. Multiple good local minima can be found for a given model and task but these minima are most often separated by regions of high loss \citep{pmlr-v119-frankle20a_lmc_lth}. Therefore, combining the parameters from multiple trained models without falling into one of these high-loss regions and destroying the features learned during training is a hard task and constitutes an active area of research.

Previous works established empirically that any two minima in an NN parameter space found through SGD and its variants are linked by a non-linear low-loss path \citep{garipov2018_fge, draxler2018_no-barriers}. The term \emph{mode connectivity} describes this phenomenon. However, to find this low-loss path between two minima one needs to run a computationally expensive optimization algorithm. As such, model fusion based on nonlinear mode connectivity has not been explored.
On the other hand, \emph{linear mode connectivity} which describes two optima connected by a \emph{linear} low loss path in the parameter space \citep{pmlr-v119-frankle20a_lmc_lth}, provides a straightforward way of combining these models. Indeed, if the loss remains low on the linear path from one model to the other, merging the two models is as simple as averaging or linearly interpolating their parameters. This has emerged as a simple, yet powerful way to compare NN minima. However, this phenomenon is very rare in practice and is not guaranteed even for networks with the same initializations \citep{pmlr-v119-frankle20a_lmc_lth}.

One reason linear mode connectivity is hard to obtain is due to the well-known NN invariance to permutations. Indeed, it is possible to permute the neurons of an NN layer without changing the actual function learned by the model as long as the connections/weights to the subsequent layer are permuted consistently. Therefore, it is possible to have the same features learned at every layer of two different NN models and them not be linearly mode-connected if the order of the features differs from one network to the other. Using this invariance as justification, \citet{entezari2022_perm-invariance-lmc} conjectured that most SGD solutions can be permuted in such a way that they are linearly mode connected to most other SGD solutions and presented empirical support for this conjecture. Many works in recent years have provided algorithms for finding permutations that successfully render pairs of SGD solutions linearly mode connected, or at least significantly lower the loss barrier on the linear path between these solutions, further supporting this conjecture \cite{tatro2020_opt-mode_con, singh-jaggi2020_OT-fusion, pena2023_sinkhorn-rebasin, ainsworth2023_git-rebasin}.

While these algorithms and the found transformations have been successful in lowering the loss barrier when interpolating between pairs of SGD solutions most of them do not consider the possibility that perhaps other linear transformations, besides permutations, would provide an even better matching of NN weights. % resulting in even better linear mode connectivity.
While the permutation conjecture is enticing given its simplicity and NNs' invariance to permutations, there is nothing inherently stopping NNs from distributing computations that are done by one neuron in a model to be done by multiple neurons in another model. Permutations would fail to capture this relationship since it is not a one-to-one mapping between features. Furthermore, the focus of recent works has been mainly on merging pairs of models, and merging multiple models has received limited study. However, if a similar function is learned by networks trained on the same task then it should be possible to extract the commonly learned features from not only two but also a larger population of models. Model merging algorithms should therefore be able to find these features and the relationship between them and then merge many models without negatively affecting performance.
\vspace{-5pt}
\paragraph{Contributions}
In this work we introduce CCA Merge, a more flexible way of merging models based on maximizing the correlation between linear combinations of neurons. Furthermore we focus on the difficult setting of merging not only two but also \emph{multiple}  models which were \emph{fully trained from random initializations}. Our main contributions are threefold:
\begin{itemize}[leftmargin=*]
    \item We propose a new model merging method based on Canonical Correlation Analysis (Sec. \ref{s:cca_merge_theory}) which we will refer to as ``CCA Merge''. This method is more flexible than past, permutation-based methods and therefore makes better use of the correlation information between neurons (Sec. \ref{ss:flexibility}).
    \item We compare CCA Merge to past works and find that it yields better performing merged models across a variety of architectures and datasets. This is true in both settings where the models were trained on the same data (Sec. \ref{ss:same_data}) or on disjoint splits of the data (Sec. \ref{ss:split_data}).
    \item We take on the difficult problem of aligning features from multiple models and then merging them. We find that CCA Merge is significantly better at finding the common learned features from populations of NNs and aligning them, leading to lesser accuracy drops as the number of models being merged increases (Sec. \ref{ss:many_models}).
\end{itemize}

% \input{sections/s2_related_work}
% \vspace{-10pt}
\section{Related Work}\label{s:related_work}
% \vspace{-5pt}
\paragraph{Mode connectivity}
\citet{freeman2017topology} proved theoretically that one-layered ReLU neural networks have asymptotically connected level sets. \citet{garipov2018_fge} and \citet{draxler2018_no-barriers} explore these ideas empirically and introduce the concept of \emph{mode connectivity} to describe ANN minima that are connected by nonlinear paths in parameter space along which the loss remains low. %, the maximum of the loss along this path was termed the \emph{energy barrier}, and both works proposed algorithms for finding such paths. %\cite{garipov2018_fge} further proposed \emph{Fast Geometric Ensembling} (FGE) as a way to take advantage of mode connectivity by ensembling multiple model checkpoints from a single training trajectory. 
\citet{pmlr-v119-frankle20a_lmc_lth} introduced the concept of \emph{linear mode connectivity} describing the scenario in which two ANN minima are connected by a \emph{linear} low loss path in parameter space. %They used linear mode connectivity to study network stability to SGD noise (i.e. different data orders and augmentations). They found that at initialization, networks are typically not stable, i.e. training with different SGD noise from a random initialization typically leads to optima that are not linearly mode connected, however early in training models become stable to such noise.
\vspace{-5pt}
\paragraph{Model merging through alignment}
More recently, \citet{entezari2022_perm-invariance-lmc} have conjectured that ``Most SGD solutions belong to a set $\mathcal{S}$ whose elements can be permuted in such a way that there is no barrier on the linear interpolation between any two permuted elements in $\mathcal{S}$'' or in other words most SGD solutions are linearly mode connected provided the right permutation is applied to align the two solutions. %They then perform experiments that support this conjecture and also empirically establish that increasing network width, decreasing depth, using more expressive architectures, or training on a simpler task all help linear mode connectivity by decreasing the loss barrier between two optima. 
Many recent works seem to support this conjecture, proposing methods for finding the ``right" permutations \citep{tatro2020_opt-mode_con, singh-jaggi2020_OT-fusion, pena2023_sinkhorn-rebasin, ainsworth2023_git-rebasin}. %For example, \cite{singh-jaggi2020_OT-fusion} and \cite{pena2023_sinkhorn-rebasin} propose optimal transport-based methods for finding the best transformation to match two models.
%The method of \cite{pena2023_sinkhorn-rebasin} is differentiable and they relax the constraint of finding a binary permutation matrix but add an entropy regularizer instead. 
%
%\cite{ainsworth2023_git-rebasin} propose an algorithm for finding the optimal permutation for merging models based on the distances between the weights of the models themselves. \cite{jordan2023repair} exposes the phenomenon in which interpolated deep networks suffer a variance collapse in their activations leading to poor performance. They propose REPAIR which mitigates this problem by rescaling the preactivations of interpolated networks through the recomputation of BatchNorm statistics. 
\vspace{-5pt}
\paragraph{``Easy'' settings for model averaging} Linear mode connectivity is hard to achieve in deep learning models. \citet{pmlr-v119-frankle20a_lmc_lth} established that even models trained on the same dataset with the same learning procedure and even the same initialization might not be linearly mode connected if they have different data orders/augmentations.
It seems that only models that are already very close in parameter space can be directly combined through linear interpolation. This is the case for snapshots of a model taken at different points during its training trajectory \citep{garipov2018_fge, izmailov2018_averaging} or multiple fine-tuned models with the same pre-trained initialization \citep{wortsman22_modelsoups, ilharco2023_task-arithmetic, yadav2023tiesmerging}. This latter setting is the one typically considered in NLP research. Another setting that is worth mentioning here is the ``federated learning'' inspired one where models are merged every couple of epochs during training \citep{mcmahan17_fedavg}. The common starting point in parameter space and the small number of training iterations before merging make LMC easier to attain. Model fusion has been very successful in these settings where aligning the models prior to merging is not required.

We emphasize that these settings are different from ours in which we aim to merge \emph{fully trained models} with different parameter initializations and SGD noise (data order and augmentations).
\vspace{-5pt}
\paragraph{Merging multiple models}
Merging more than two models has only been explored thoroughly in the ``easy'' settings described above. For example, \citet{wortsman22_modelsoups} average models fine-tuned with different hyperparameter configurations and find that this improves accuracy and robustness. \citet{jolicoeurmartineau2023_papa} average the weights of a population of neural networks multiple times during training, leading to performance gains.
On the other hand, works that have focused on providing feature alignment methods to be able to merge models in settings in which LMC is not trivial have mainly done so for 2 models at the time
%or for populations of simpler models such as MLPs trained on MNIST
\citep{singh-jaggi2020_OT-fusion, ainsworth2023_git-rebasin, pena2023_sinkhorn-rebasin, jordan2023repair}. An exception to this is Git Re-Basin \citep{ainsworth2023_git-rebasin} which proposes a ``Merge Many'' algorithm for merging a set of multiple models by successively aligning each model to the average of all the other models. However, results obtained with this method, which they use to merge up to 32 models, only concern the very simple set-up of MLPs trained on MNIST. \citet{singh-jaggi2020_OT-fusion} also consider merging multiple models but either in a similarly simple set-up, i.e. 4 MLPs trained on MNIST, or they fine-tune the resulting model after merging up to 8 VGG11 models trained on CIFAR100.
We extend this line of work to more challenging settings, using more complex model architectures, we report the merged models accuracies directly without fine-tuning and make this a key focus in our work.
\vspace{-5pt}
\paragraph{Model merging beyond permutations} We note that the two model merging methods based on optimal transport \cite{singh-jaggi2020_OT-fusion, pena2023_sinkhorn-rebasin} can also align models beyond permutations. However, in \citet{singh-jaggi2020_OT-fusion} this only happens when the two models being merged have different numbers of neurons at each layer. When the models have the same number of neurons the alignment matrix found by their method is a permutation, as such the majority of their results are with permutations. The method proposed by \citet{pena2023_sinkhorn-rebasin} isn't constrained to finding binary permutation matrices but binarity is still encouraged through the addition of an entropy regularizer. Furthermore, our CCA based method is different in nature from both of these since it is not inspired by optimal transport theory.
\vspace{-5pt}
\paragraph{CCA in deep learning}
In the general context of deep learning, Canonical Correlation Analysis has been used to align and compare learned representations in deep learning models \cite{raghu2017_svcca, morcos2018_insights, gotmare2018a}, a task which is similar to the feature matching conducted by model merging algorithms. These past works serve as great motivation for the present paper.
\vspace{-2pt}
\section{Using CCA to Merge Models}\label{s:cca_merge_theory}
\vspace{-3pt}
\subsection{Merging Models: Problem Definition}
\vspace{-5pt}
Let $\mathcal{M}$ denote a standard multilayer perceptron (MLP) and layer $L_i\in\mathcal{M}$ denote a linear layer of that model with a $\sigma=\text{ReLU}$ activation function, weights $\mathbf{W}_i\in\mathbb{R}^{n_i\times n_{i-1}}$ and bias $\mathbf{b}_i\in\mathbb{R}^{n_i}$. Its input is the vector of embeddings from the previous layer $\mathbf{x}_{i-1}\in\mathbb{R}^{n_{i-1}}$ and its output can be described as:
\[\mathbf{x}_i = \sigma(\mathbf{W}_i\cdot\mathbf{x}_{i-1}+\mathbf{b}_i)\]
We use `` $\cdot$ '' to denote standard matrix multiplication. 
%Let $\{L_i^\mathcal{M}\}_{i=1}^N$ be the set of layers of model $\mathcal{M}\in\{\mathcal{A}, \mathcal{B}\}$ and let $\mathbf{X}_i^\mathcal{M}\in\mathbb{R}^{m\times n_i}$ denote the set of outputs of the $i$-th layer of model $\mathcal{M}$ in response to $m$ given inputs. 
% Given the context, we assume $\mathbf{X}_i^\mathcal{M}$ is centered so that each column has a mean of 0. 

\paragraph{Problem statement}
Now consider two deep learning models $\mathcal{A}$ and $\mathcal{B}$ with the same architecture.
We are interested in the problem of merging the parameters from models $\mathcal{A}$ and $\mathcal{B}$ in a layer-by-layer fashion. As mentioned in Sec. \ref{s:related_work}, simply interpolating the models parameters typically doesn't work when the models are trained from scratch from different initializations. We therefore need to \emph{align} the two models' features before averaging them. We use the term ``feature” in our work to refer to the individual outputs of a hidden layer neuron within a neural network. We sometimes also use the term ``neuron” to refer to its learned feature or, vice-versa, we might use the term ``feature” to refer to a neuron and its parameters.
Mathematically, we are looking for linear transformations $\mathbf{T}_i\in\mathbb{R}^{n_i\times n_i}$ which can be applied at the output level of model $\mathcal{B}$ layer $i$ parameters to maximize the alignment with model $\mathcal{A}$'s parameters and minimize the interpolation error. The output of the transformed layer $i$ of $\mathcal{B}$ is then:
\[\mathbf{x}_i^\mathcal{B} = \sigma(\mathbf{T}_i\cdot\mathbf{W}_i^\mathcal{B}\cdot\mathbf{x}_{i-1}^\mathcal{B}+\mathbf{T}_i\cdot\mathbf{b}^\mathcal{B}_i)\]
We therefore also need to apply the inverse of $\mathbf{T}_i$ at the input level of the following layer's weights to keep the flow of information consistent inside a given model:
\[\mathbf{x}_{i+1}^\mathcal{B} = \sigma(\mathbf{W}_{i+1}^\mathcal{B}\cdot\mathbf{T}_i^{-1}\cdot\mathbf{x}_{i}^\mathcal{B}+\mathbf{b}^\mathcal{B}_{i+1})\]
After finding transformations $\{\mathbf{T}_i\}_{i=1}$ for every merging layer in the network we can merge the two model's parameters at every layer:
\begin{equation}\label{eq:merging}
    \mathbf{W}_i = \frac{1}{2}(\mathbf{W}_i^\mathcal{A} + \mathbf{T}_i\cdot\mathbf{W}_i^\mathcal{B}\cdot\mathbf{T}^{-1}_{i-1})
    %;\quad \mathbf{b}_i=\frac{1}{2}(\mathbf{b}_i^\mathcal{A} + \mathbf{T}_i \cdot \mathbf{b}_i^\mathcal{B})
\end{equation}

For the biases we have $\mathbf{b}_i=\frac{1}{2}(\mathbf{b}_i^\mathcal{A} + \mathbf{T}_i \cdot \mathbf{b}_i^\mathcal{B})$. The linear transformations $\mathbf{T}$ therefore need to be invertible so we can undo their changes at the input level of the next layer and they need to properly align the layers of models $\mathcal{A}$ and $\mathcal{B}$. This problem statement is a generalization of the one considered in past works where transformations $\mathbf{T}$ were constrained to being permutations \citep{tatro2020_opt-mode_con, entezari2022_perm-invariance-lmc, ainsworth2023_git-rebasin, jordan2023repair}.

We note that while artificial neural networks are invariant to permutations in the order of their neurons, this is not the case for general invertible linear transformations. Therefore, after applying the transformations to model $\mathcal{B}$ to align it to $\mathcal{A}$ its functionality might be altered. However, our results (Sec. \ref{s:results}) seem to suggest that the added flexibility of merging linear combinations of features outweighs the possible negative effects of this loss in functionality.

\paragraph{Practical considerations}
In practice, it is often easier to keep model $\mathcal{A}$ fixed and to find a way to transform model $\mathcal{B}$ such that the average of their weights can yield good performance, as opposed to transforming the parameters of both models.
Also, depending on the model architecture, it might not be necessary to compute transformations after each single layer, for example, skip connections preserve the representation space, and the last layers of models are already aligned by the training labels. Therefore we refer to the specific layers in a network where transformations must be computed as ``merging layers''. 

\paragraph{Merging multiple models} In the case where multiple models are being merged there is a simple way of extending any method which aligns features between two models to the multiple models scenario. Suppose we have a set of models $\{\mathcal{M}_i\}_{i=1}^n$ which we want to merge. We can pick one of them, say $\mathcal{M}_j$ for $1\leq j\leq n$, to be the \emph{reference model}. Then we can align the features of every other model in the set to those of the reference model and average the weights. While this ``all-to-one'' merging approach is quite straightforward it seems to work well in practice.

% \subsection{Background on Canonical Correlation Analysis}
\subsection{CCA Merge: Merging models with CCA}
The ``best'' way to align two layers from two different deep learning models and compute the transformations $\mathbf{T}$ is still an open-question and has been the main differentiating factor between past works (see the baselines presented in Sec. \ref{ss:same_data} and ``Model merging through alignment'' in Sec. \ref{s:related_work}).

We propose the use of Canonical Correlation Analysis (CCA) to find the transformations which maximize the correlations between linear combinations of the original features from models $\mathcal{A}$ and $\mathcal{B}$. 
Let $\mathbf{X}_i^\mathcal{M}\in\mathbb{R}^{m\times n_i}$ denote the set of outputs (internal representations or neural activations) of the $i$-th layer of model $\mathcal{M}\in\{\mathcal{A}, \mathcal{B}\}$ in response to $m$ given input examples. We center $\mathbf{X}_i^\mathcal{M}$ so that each column (feature or neuron) has a mean of 0.

CCA finds projection matrices $\mathbf{P}_i^\mathcal{A}$ and $\mathbf{P}_i^\mathcal{B}$ which bring the neural activations $\mathbf{X}_i^\mathcal{A}$ and $\mathbf{X}_i^\mathcal{B}$ respectively from their original representation spaces into a new, common, representation space through the multiplications $\mathbf{X}_i^\mathcal{A}\cdot \mathbf{P}_i^\mathcal{A}$ and $\mathbf{X}_i^\mathcal{B}\cdot \mathbf{P}_i^\mathcal{B}$. The features of this new representation space are orthogonal linear combinations of the original features of $\mathbf{X}_i^\mathcal{A}$ and $\mathbf{X}_i^\mathcal{B}$, and they maximize the correlations between the two projected sets of representations.

% To merge models using CCA we simply apply the CCA algorithm to the two sets of activations $X_i^\mathcal{A}$ and $X_i^\mathcal{B}$ to get the corresponding projection matrices $P_i^\mathcal{A}$ and $P_i^\mathcal{B}$.

Once the two projection matrices $\mathbf{P}_i^\mathcal{A}$ and $\mathbf{P}_i^\mathcal{B}$ aligning $\mathbf{X}_i^\mathcal{A}$ and $\mathbf{X}_i^\mathcal{B}$ respectively have been found through CCA, we can define the transformation $\mathbf{T}_i=\nolinebreak \left(\mathbf{P}_i^\mathcal{B}\cdot{\mathbf{P}_i^\mathcal{A}}^{-1}\right)^\top$. This transformation can be thought of as first bringing the activations of model $\mathcal{B}$ into the common, maximally correlated space between the two models by multiplying by $\mathbf{P}_i^\mathcal{B}$ and then applying the inverse of $\mathbf{P}_i^\mathcal{A}$ to go from the common space found by CCA to the embedding space of model $\mathcal{A}$. 
%The reason for doing this instead of transforming both the weights of model A and model B with $\mathbf{P}^\mathcal{A}$ and $\mathbf{P}^\mathcal{B}$ is because of the ReLU non-linearity. The common representation space found by CCA has no notion of directionality, and might contain important information even in negative orthants (high-dimensional quadrants where at least one of the variables has negative values) that might get squashed to zero by ReLU non-linearities. The representation space of model A doesn't have this problem. 
The transpose here is simply to account for the fact that $\mathbf{T}_i$ multiplies $\mathbf{W}_i$ on the left while the $\mathbf{P}^\mathcal{M}_i$'s were described as multiplying $\mathbf{X}^\mathcal{M}_i$ on the right, for $\mathcal{M}\in\{\mathcal{A}, \mathcal{B}\}$. The averaging of the parameters of model $\mathcal{A}$ and transformed $\mathcal{B}$ can then be conducted following Eq.~\ref{eq:merging}.

\paragraph{Background on CCA}
 % \textcolor{blue}{For the following we omit the layer index $i$ since it is implicit.}
In this section we omit the layer index $i$ since it is implicit. CCA finds the projection matrices $\mathbf{P}^\mathcal{A}$ and $\mathbf{P}^\mathcal{B}$ by iteratively defining vectors $\mathbf{p}^\mathcal{A}$ and $\mathbf{p}^\mathcal{B}$ in $\mathbb{R}^{n}$ such that the projections $\mathbf{X}^\mathcal{A}\cdot\mathbf{p}^\mathcal{A}$ and $\mathbf{X}^\mathcal{B}\cdot\mathbf{p}^\mathcal{B}$ have maximal correlation and norm 1.

Let $\mathbf{S}^{\mathcal{A}\mathcal{A}}=(\mathbf{X}^\mathcal{A})^\top\cdot \mathbf{X}^\mathcal{A}$, $\mathbf{S}^{\mathcal{B}\mathcal{B}}=(\mathbf{X}^\mathcal{B})^{\top}\cdot \mathbf{X}^\mathcal{B}$ and $\mathbf{S}^{\mathcal{A}\mathcal{B}}=\nolinebreak(\mathbf{X}^\mathcal{A})^\top\cdot\mathbf{X}^\mathcal{B}$ denote the scatter matrix of $\mathbf{X}^\mathcal{A}$, the scatter matrix of $\mathbf{X}^\mathcal{B}$ and the cross-scatter matrix of $\mathbf{X}^\mathcal{A}$ and $\mathbf{X}^\mathcal{B}$ respectively.
%In practice, it is possible to use the respective covariance matrix which is simply $\mathbf{C}=\frac{1}{n-1}\mathbf{S}$ as it will yield the same solution vectors up to multiplication by a constant. The main optimization objective of CCA can be written as:
%
\begin{align*}
    \mathbf{p}^\mathcal{A}, \mathbf{p}^\mathcal{B} &= \argmax\limits_{\mathbf{p}^\mathcal{A}, \mathbf{p}^\mathcal{B}} (\mathbf{p}^\mathcal{A})^\top \mathbf{S}^{\mathcal{A}\mathcal{B}} \mathbf{p}^\mathcal{B}\\
    \text{s.t.} &\quad \|\mathbf{X}^\mathcal{A}\cdot\mathbf{p}^\mathcal{A}\|^2 = (\mathbf{p}^\mathcal{A})^\top\cdot \mathbf{S}^{\mathcal{A}\mathcal{A}} \cdot\mathbf{p}^\mathcal{A} = 1\\
    & \quad \|\mathbf{X}^\mathcal{B}\cdot\mathbf{p}^\mathcal{B}\|^2 = (\mathbf{p}^\mathcal{B})^\top\cdot \mathbf{S}^{\mathcal{B}\mathcal{B}} \cdot\mathbf{p}^\mathcal{B} = 1
\end{align*}
Since these vectors $\mathbf{p}^\mathcal{A}$ and $\mathbf{p}^\mathcal{B}$ are defined iteratively they are also required to be orthogonal to the vectors found previously in the metrics defined by $\mathbf{S}^{\mathcal{A}\mathcal{A}}$ and $\mathbf{S}^{\mathcal{B}\mathcal{B}}$ respectively. Formulating this as an ordinary eigenvalue problem and making it symmetric by a change of variables allows us to find the closed-form solutions as being the vectors in $\mathbf{P}^\mathcal{A}$ and $\mathbf{P}^\mathcal{B}$ defined by:
% This can be formulated as an ordinary eigenvalue problem:
% %
% \begin{align*}
%     \begin{pmatrix}
%         \mathbf{0} & S_{XX}^{-1}S_{XY}\\
%         S_{YY}^{-1}S_{YX} & \mathbf{0}
%     \end{pmatrix}
%     \begin{pmatrix}
%         p_X\\
%         p_Y
%     \end{pmatrix} = \lambda
%     \begin{pmatrix}
%         p_X\\
%         p_Y
%     \end{pmatrix}
% \end{align*}
% %
% Which can be made symmetric by introducing the vectors $v_X = S_{XX}^{1/2}p_X$ and $v_Y = S_{YY}^{1/2}p_Y$. 
% %
% \begin{align*}
%     \begin{pmatrix}
%         \mathbf{0} & S_{XX}^{-1/2}S_{XY}S_{YY}^{-1/2}\\
%         S_{YY}^{-1/2}S_{YX}S_{XX}^{-1/2} & \mathbf{0}
%     \end{pmatrix}
%     \begin{pmatrix}
%         v_X\\
%         v_Y
%     \end{pmatrix} = \lambda
%     \begin{pmatrix}
%         v_X\\
%         v_Y
%     \end{pmatrix}
% \end{align*}
% %
% We can then find $v_X$ and $v_Y$ as being the left and right singular vectors of $S_{XX}^{-1/2}S_{XY}S_{YY}^{-1/2}$ respectively. The projection vectors are therefore $p_X = S_{XX}^{-1/2}v_X$ and $p_Y = S_{YY}^{-1/2}v_Y$. If we want the dimensionality of the common space to be the same as the embedding spaces we can use the whole set of left and right singular vectors of $S_{XX}^{-1/2}S_{XY}S_{YY}^{-1/2}$:
%
\begin{align*}
    \mathbf{U}, \mathbf{S}, \mathbf{V}^\top &= \text{SVD}((\mathbf{S}^{\mathcal{A}\mathcal{A}})^{-1/2}\cdot\mathbf{S}^{\mathcal{A}\mathcal{B}}\cdot(\mathbf{S}^{\mathcal{B}\mathcal{B}})^{-1/2})\\
    \mathbf{P}^\mathcal{A} &= (\mathbf{S}^{\mathcal{A}\mathcal{A}})^{-1/2}\cdot\mathbf{U} \quad \text{and} \quad \mathbf{P}^\mathcal{B} = (\mathbf{S}^{\mathcal{B}\mathcal{B}})^{-1/2}\cdot\mathbf{V} 
\end{align*}
In practice we use Regularized CCA to make the computation of $\mathbf{P}^\mathcal{A}$ and $\mathbf{P}^\mathcal{B}$ more robust (see App. \ref{a:reg_cca}).
%where we add $\gamma\mathbf{I}$ to both scatter matrices $\mathbf{S}^{\mathcal{A}\mathcal{A}}$ and $\mathbf{S}^{\mathcal{B}\mathcal{B}}$}
For more details we direct the reader to \citet{DeBie2005} from which this section was inspired. 

% \subsection{CCA Merge: Merging models with CCA}
% To merge models using CCA we simply apply the CCA algorithm to the two sets of activations $X_i^\mathcal{A}$ and $X_i^\mathcal{B}$ to get the corresponding projection matrices $P_i^\mathcal{A}$ and $P_i^\mathcal{B}$. Using the framework described above of matching model $\mathcal{B}$ to model $\mathcal{A}$, which is consistent with past works, we can define $T_i=\nolinebreak \left(P_i^\mathcal{B}{P_i^\mathcal{A}}^{-1}\right)^\top$. The transpose here is to account for the fact that $T_i$ multiplies $W_i$ on the left while the $P_i$s were described as multiplying $X_i$ on the right. This transformation can be thought of as first bringing the activations of model $\mathcal{B}$ into the common, maximally correlated space between the two models by multiplying by $P_i^\mathcal{B}$ and then applying the inverse of $P_i^\mathcal{A}$ to go from the common space to the embedding space of $\mathcal{A}$. The averaging of the parameters of model $\mathcal{A}$ and transformed $\mathcal{B}$ can then be conducted following Eq.~\ref{eq:merging}.

\vspace{-5pt}
\section{Results}\label{s:results}
\vspace{-3pt}
\subsection{Experimental Details}
\vspace{-5pt}
We trained VGG11 models \citep{simonyan2015a_vgg} on CIFAR10 \citep{krizhevsky2009_cifar}, ResNet20 models on CIFAR100 and ResNet18 models on ImageNet \citep{russakovsky2015_imagenet}. %(full-sized images but training and evaluating only on 200 of the 1k classes). %using the one-hot encodings of the labels as training objectives
We trained models of different widths, multiplying their original width by $w\in\{1, 2, 4, 8\}$. %In addition we also
The models were trained either using the one-hot encodings of the labels or the CLIP \citep{radford2021_clip} embeddings of the class names as training objectives. This last setting is similar to the one used by \citet{stoica2024zipit} and we found it to yield better learned representations and performances on the task as well as less variability between random initializations.

\subsection{CCA's flexibility allows it to better model relations between neurons}\label{ss:flexibility}
We first aim to illustrate the limits of permutation based matching and the flexibility offered by CCA Merge.
Suppose we want to merge two models, $\mathcal{A}$ and $\mathcal{B}$, at a specific merging layer, and let $\{\mathbf{z}_i^\mathcal{M}\}_{i=1}^n$ denote the set of neurons of model $\mathcal{M}\in\{\mathcal{A}, \mathcal{B}\}$ at that layer. We note here that, in terms of network weights, $\mathbf{z}_i^\mathcal{M}$ simply refers to the $i$-th row of the weight matrix $\mathbf{W}^\mathcal{M}$ at that layer. Given the activations of the two sets of neurons in response to a set of given inputs, we can compute the correlation matrix $\mathbf{C}$ where element $\mathbf{C}_{ij}$ is the correlation between neurons $\mathbf{z}_i^\mathcal{A}$ and $\mathbf{z}_j^\mathcal{B}$. For each neuron $\mathbf{z}_i^\mathcal{A}$, for $1\leq i\leq n$, the distribution of its correlations with all neurons from model $\mathcal{B}$ is of key interest for the problem of model merging. If, as the permutation hypothesis implies, there exists a one-to-one mapping between $\{\mathbf{z}_i^\mathcal{A}\}_{i=1}^n$ and $\{\mathbf{z}_i^\mathcal{B}\}_{i=1}^n$, then we would expect to have one highly correlated neuron for each $\mathbf{z}_i^\mathcal{A}$ -- say $z_j^\mathcal{B}$ for some $1\leq j\leq n$ -- and all other correlations $\mathbf{C}_{ik}$, $k\neq j$, close to zero. 
On the other hand, if there are multiple neurons from model $\mathcal{B}$ highly correlated with $z_i^\mathcal{A}$, this would indicate that the feature learned by $z_i^\mathcal{A}$ is distributed across multiple neurons in model $\mathcal{B}$ -- a relationship that CCA Merge would capture.

\begin{figure}[ht]
\begin{center}
\includegraphics[width=0.475\textwidth]{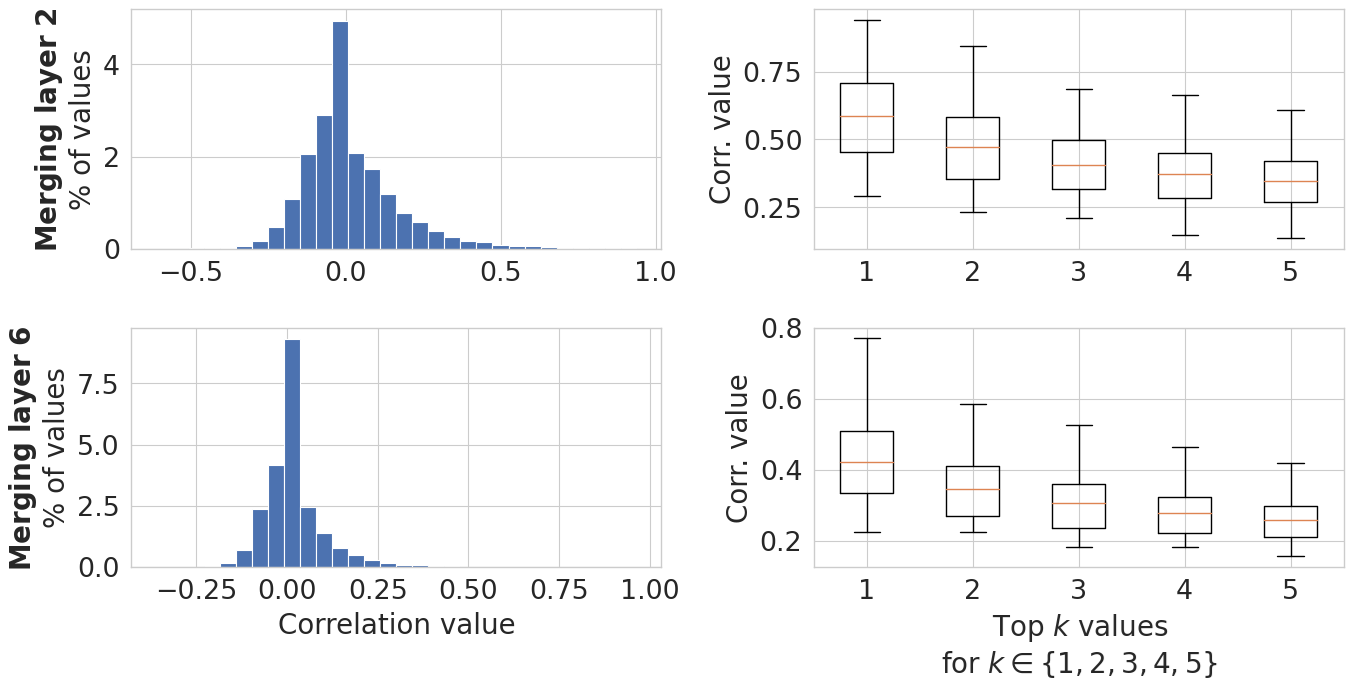}
\end{center}
\vspace{-15pt}
\caption{\textbf{Left column:} distribution of correlation values between the neurons $\{\mathbf{z}_i^\mathcal{A}\}_{i=1}^n$ and $\{\mathbf{z}_i^\mathcal{B}\}_{i=1}^n$ of two ResNet20x8 models ($\mathcal{A}$ and $\mathcal{B}$) trained on CIFAR100 at two different merging layers; \textbf{Right column:} for $k\in\{1,2,3,4,5\}$ the distributions of the top $k$-th correlation values for all neurons in model $\mathcal{A}$ at those merging layers.}
\vspace{-7pt}
\label{fig:corr_dists}
\end{figure}

In the left column of Fig.~\ref{fig:corr_dists}, we plot the distributions of the correlations between two ResNet20x8 models (i.e., all the elements from the correlation matrix $\mathbf{C}$) for 2 different merging layers. 
%, one at the beginning, one in the middle and one at the end of the models. 
The vast majority of correlations have values around zero, as expected, since each layer learns multiple different features. 
In the right column of Fig. \ref{fig:corr_dists} we use box plots to show the values of the top 5 correlation values across all $\{\mathbf{z}_i^\mathcal{A}\}_{i=1}^n$. 
For each neuron $\mathbf{z}_i^\mathcal{A}$, we select its top $k$-th correlation from $\mathbf{C}$ and we plot these values for all neurons $\{\mathbf{z}_i^\mathcal{A}\}_{i=1}^n$. For example, for $k=1$, we take the value $\max\limits_{1\leq j\leq n} \mathbf{C}_{ij}$, for $k=2$ we take the second largest value from the $i$-th row of $\mathbf{C}$, and so on. We observe the top correlations values are all relatively high but none of them approaches full correlation (i.e., value of one), suggesting that the feature learned by each neuron $\mathbf{z}_i^\mathcal{A}$ from model $\mathcal{A}$ is distributed across multiple neurons from $\mathcal{B}$ -- namely, those having high correlations -- as opposed to having a single highly correlated match.

\begin{figure}[ht]
\begin{center}
\includegraphics[width=0.465\textwidth]{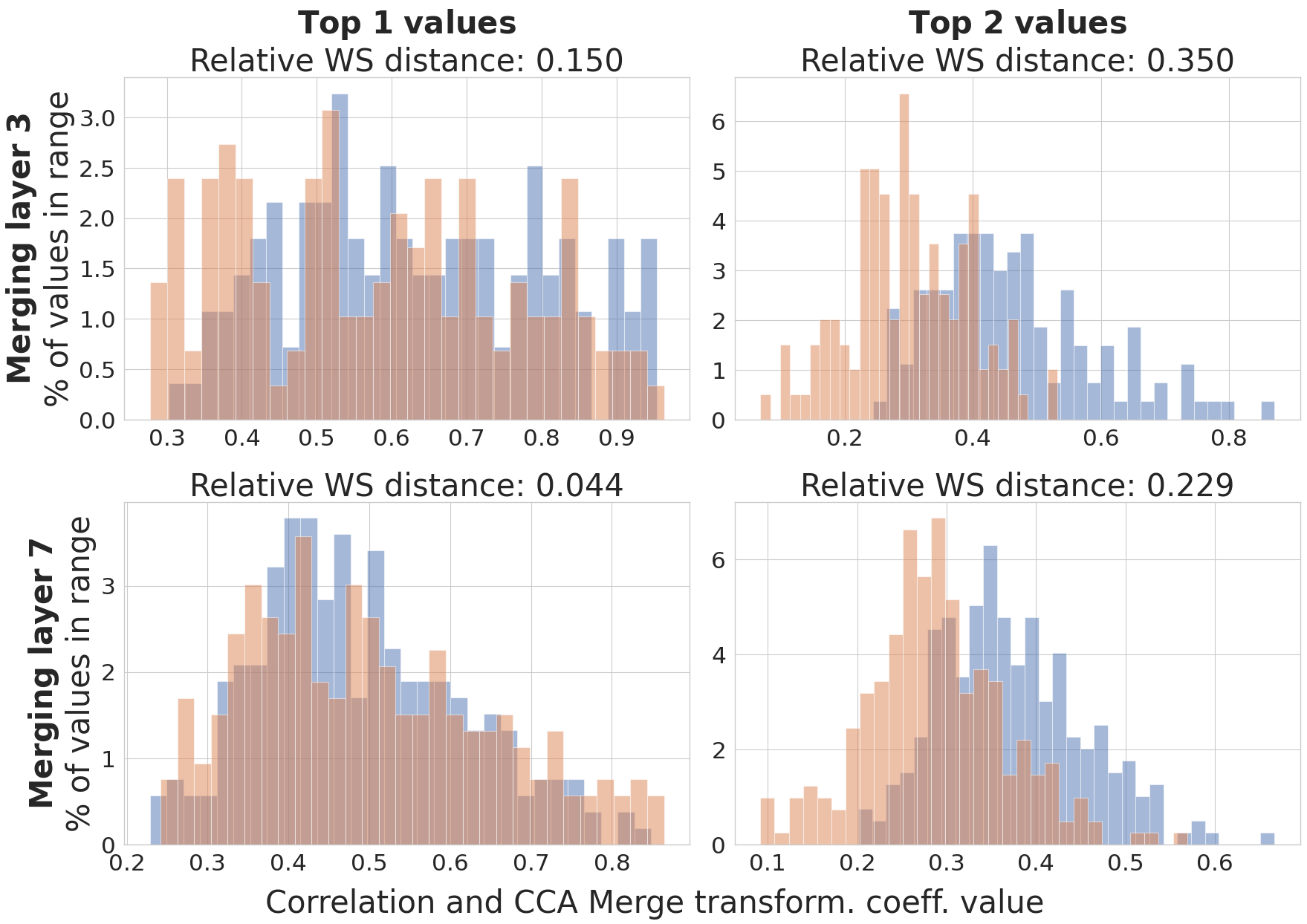}
\end{center}
\vspace{-12pt}
\caption{Distributions of top 1 (\textbf{left column}) and 2 (\textbf{right column}) correlations (blue) and CCA Merge transformation coefficients (orange) across neurons from model $\mathcal{A}$ at two different merging layers. In the left column for example, for each neuron $\mathbf{z}_i^\mathcal{A}$ we have one correlation value corresponding to $\max_{1\leq j\leq n} \mathbf{C}_{ij}$ and one coefficient value corresponding to $\max_{1\leq j\leq n} \mathbf{T}_{ij}$ where $\mathbf{C}$ is the cross-correlation matrix between neurons of models $\mathcal{A}$ and $\mathcal{B}$, and $\mathbf{T}$ is the CCA Merge transformation matching neurons of $\mathcal{B}$ to those of $\mathcal{A}$. Wasserstein distance between the distributions of top $k\in\{1,2\}$ correlations and top $k$ Merge CCA coefficients are reported, relative to equivalent distances between correlations and Permute transforms (all top 1 values are 1, and top 2 values are 0).}
\vspace{-10pt}
\label{fig:cca_corrs}
\end{figure}

Given the flexibility of CCA Merge, we expect it to better capture these relationships between the neurons of the two networks. 
We recall that CCA Merge computes a linear transformation $\mathbf{T}$
that matches to each neuron $\mathbf{z}_i^\mathcal{A}$ a linear combination $\mathbf{z}_i^\mathcal{A} \approx \sum_{j=1}^n \mathbf{T}_{ij}\mathbf{z}_j^\mathcal{B}$ of the neurons in $\mathcal{B}$.
We expect the distribution of the coefficients (i.e., elements of $\mathbf{T}$) to match the distribution of the correlations ($\mathbf{C}_{ij}$ elements), indicating the linear transformation found by CCA Merge adequately models the correlations and relationships between the neurons of the two models. For each neuron $\mathbf{z}_i^\mathcal{A}$, we select its top $k$-th, for $k\in\{1, 2\}$, correlation from the $i$-th row of $\mathbf{C}$ and its top $k$-th coefficient from the $i$-th row of $\mathbf{T}$ and we plot a histogram of these values for all neurons $\{\mathbf{z}_i^\mathcal{A}\}_{i=1}^n$ in Fig.~\ref{fig:cca_corrs}. Indeed, the distributions of the correlations and those of the CCA Merge coefficients are visually similar, albeit not fully coinciding. To quantify this similarity we compute the Wasserstein distance between these distributions, normalized by the equivalent quantity if the transformation was a permutation matrix. For a permutation matrix, the top 1 values would be of 1 for every neuron $\mathbf{z}_i^\mathcal{A}$ and all other values would be 0.
We can see that CCA Merge finds coefficients that closely match the distribution of the correlations, more so than simple permutations, since the ratio of the two distances are 0.15 and 0.04, respectively, for top 1 values in the two considered layers and 0.35 and 0.23 for top 2 values. 
%Top 2 relative distances are a bit higher at 0.35, 0.23 and 0.28 respectively since the correlations distribution is closer to zero than to one therefore closer to top 2 permutation coefficients, however it is still the case that the CCA Merge coefficients have a distribution more closely matching that of the correlations.

Furthermore, when using permutations to merge 2 models, $\mathcal{A}$ and $\mathcal{B}$, a large percentage (25-50\%) of the neurons from model $\mathcal{A}$ do not get matched with their highest correlated neuron from model $\mathcal{B}$ by the permutation matrix, we call these non-optimal matches. In such cases, the relationship between these highly correlated but not matched neurons is completely ignored during merging. CCA Merge on the other hand consistently assigns high transformation coefficients to the top correlated neurons. See App. \ref{a:non_opt_matches_permute} for more details.

\subsection{Models merged with CCA Merge achieve better performance}\label{ss:same_data}
In Table \ref{t:results_same_data} we show the test accuracies of merged VGG11 and ResNet20 models of different widths trained on CIFAR10 and CIFAR100 respectively for CCA Merge and multiple other popular model merging methods. The number of models being merged is 2 and for each experiment, we report the mean and standard deviation across 4 merges where the base models were trained with different initialization, data augmentation, and data order seeds. Results for more widths can be found in App. \ref{a:same_data}.
We report the average accuracies of the base models being merged under the label ``Base models avg.'' (i.e. each model is evaluated individually and their accuracies are then averaged) as well as the accuracies of ensembling the models (the logits of the different models are averaged and the final prediction is the argmax). Ensembling is considered to be the upper limit of what model fusion methods can achieve. Also, since the models being merged were trained on the same data, we do not expect the merged models to outperform the endpoint ones in this particular setting, see App. \ref{a:merged_vs_endpoints} for more details.
We compare CCA Merge with the following methods:
\vspace{-5pt}
\begin{itemize}[leftmargin=*]
\setlength\itemsep{0em}
    \item \textbf{Direct averaging:} averaging the models' weights without applying any transformation to align neurons. This is equivalent to $\mathbf{T}=\mathbf{T^{-1}}=\mathbf{I}$, the identity matrix.
    \item \textbf{Permute:} permuting model weights to align them, the permutation matrix is found by solving the linear sum assignment problem consisting of maximizing the sum of correlations between matched neurons. This method is equivalent to the ``Matching Activations'' one from \citet{ainsworth2023_git-rebasin}, the ``Permute'' method considered in \citet{stoica2024zipit} and the neuron alignment algorithm proposed by \citet{li2015_convergent, tatro2020_opt-mode_con} and used in \citet{jordan2023repair}.
    \item \textbf{OT Fusion:} Using optimal transport to align neurons. This is the method presented in \citet{singh-jaggi2020_OT-fusion}.
    \item \textbf{Matching Weights:} permuting model weights by directly minimizing the distance between the two model weights by solving a sum of bilinear assignments problem (SOBLAP). This is the main method from \citet{ainsworth2023_git-rebasin}.
    \item \textbf{ZipIt!:} model merging method proposed by \citet{stoica2024zipit}. We note that ZipIt! also allows the merging of neurons from the same network which results in a redundancy reduction effect that the other methods do not have. Also, it isn't strictly speaking a permutation-based method although similar.
\end{itemize}
%
% For ResNets, we apply REPAIR \citep{jordan2023repair} to recompute the BatchNorm statistics after the weight averaging and before evaluating the merged model to avoid variance collapse.
The models were trained from scratch from different initializations, the merging was based on the training data (to compute activation statistics) but all accuracies reported are on the test set. For ResNets, we recompute the BatchNorm statistics after the weight averaging and before evaluation as suggested by \citet{jordan2023repair} to avoid variance collapse.

\begin{table*}[ht]
    \centering
    \begin{tabular}{l|l l|l l}
        & VGG11$\times 1$ & VGG11$\times 8$ & ResNet20$\times 1$ & ResNet20$\times 8$\\
        Method & CIFAR10 & CIFAR10 & CIFAR100 & CIFAR100\\
        \noalign{\smallskip}
        \hline
        \noalign{\smallskip}
        \textcolor{gray}{Base models avg.} & \textcolor{gray}{87.27 \small{$\pm$0.25\%}} & \textcolor{gray}{88.20 \small{$\pm$0.45\%}} & \textcolor{gray}{69.21 \small{$\pm$0.22\%}} & \textcolor{gray}{78.77 \small{$\pm$0.28\%}}\\
        \textcolor{gray}{Ensemble}         & \textcolor{gray}{89.65 \small{$\pm$0.13\%}} & \textcolor{gray}{90.21 \small{$\pm$0.24\%}}  & \textcolor{gray}{73.51 \small{$\pm$0.20\%}} & \textcolor{gray}{80.98 \small{$\pm$0.21\%}}\\
        \noalign{\smallskip}
        \hline
        \noalign{\smallskip}
        Direct averaging & 10.54 \small{$\pm$0.93\%} & 10.45 \small{$\pm$0.74\%}  & 1.61 \small{$\pm$0.16\%}  & 14.00 \small{$\pm$1.66\%}\\
        Permute          & 54.39 \small{$\pm$6.45\%} & 62.58 \small{$\pm$3.31\%}  & 28.76 \small{$\pm$2.20\%} & 72.90 \small{$\pm$0.08\%}\\
        OT Fusion          & 53.86 \small{$\pm$10.4\%} & 68.32 \small{$\pm$3.13\%}  & 29.05 \small{$\pm$2.55\%} & 72.45 \small{$\pm$0.08\%}\\
        Matching Weights  & 55.40 \small{$\pm$4.67\%} & 73.74 \small{$\pm$1.77\%}  & 21.38 \small{$\pm$4.36\%} & 74.29 \small{$\pm$0.51\%}\\
        ZipIt!           & 52.93 \small{$\pm$6.37\%} &          -          & 25.26 \small{$\pm$2.30\%} & 72.47 \small{$\pm$0.41\%}\\
        \rowcolor{gray!25}\textbf{CCA Merge (ours)}& \textbf{82.65 \small{$\pm$0.73\%}} & \textbf{84.36 \small{$\pm$2.09\%}} & \textbf{31.79 \small{$\pm$1.97\%}} & \textbf{75.06 \small{$\pm$0.18\%}} \\
    \end{tabular}
    \caption{VGG11 trained on CIFAR10 \& ResNet20 trained on CIFAR100 - Accuracies and standard deviations from 4 different merges of 2 models are presented. Models averaged with CCA Merge notably outperform models merged with other methods, narrowing the gap between merged models and model ensembles. Model ensembles are significantly more memory and compute expensive and represent the upper bound of attainable performance for model merging methods.}
    % \vspace{-5pt}
    \label{t:results_same_data}
    % \vspace{-5pt}
\end{table*}

VGG11 models merged with CCA Merge have significantly higher accuracies than models merged with any other method, and this is true across all model widths considered. Differences in accuracy ranging from 10\% ($\times 8$ width) up to 25\% ($\times 1$ width) can be observed between CCA Merge and the second-best performing method. Furthermore, CCA Merge is more robust when merging smaller width models, incurring smaller accuracy drops than other methods when the width is decreased from $\times8$ to $\times1$; 1.71\% drop for CCA Merge versus 18.34\% for Matching Weights and 8.19\% for Permute. Lastly, CCA Merge seems to be more stable across different initializations, the accuracies having smaller standard deviations than all other methods for the same width except for Matching Weights for $\times 8$ width models.
We note that for VGG models with width multipliers above $\times 2$, we ran into out-of-memory issues when running ZipIt!, which is why those results are not present. The same conclusions seem to hold for ResNets20 trained on CIFAR100, although the differences in performance here are less pronounced. 
% Here we only compare with ZipIt! and Permute since \cite{jordan2023repair} and \cite{stoica2023zipit} have suggested that Permute is better than Matching Weights in this setting. Across all model widths, we observe that CCA Merge yields merged models having better accuracies than other methods. % Furthermore, our method seems again to be more stable across different random seeds, having lower standard deviations than other methods.

In Table \ref{t:results_imagenet} we present the performance of merged ResNet18 models of width 4 trained on ImageNet. In this setting we ran into OOM issues with ZipIt! therefore we only compare with Direct averaging, Permute, OT Fusion and Matching Weights, the four of which are significantly outperformed by CCA Merge. CCA Merge reduces the gap between model merging methods and model ensembles.
%Furthermore, the top 5 accuracy of models merged with CCA Merge is remarkably close to the accuracy of model ensembles, the peak of attainable performance.

% Table \ref{t:resnet20_cifar100} contains similar results but for ResNet20 trained on CIFAR100. Here we only compare with ZipIt! and Permute since \cite{jordan2023repair} and \cite{stoica2023zipit} have suggested that Permute is better than Matching Weights in this setting. Again, across all model widths, we observe that CCA Merge yields merged models having better accuracies than other methods, although the differences here are less pronounced. Furthermore, our method seems again to be more stable across different random seeds, having lower standard deviations than other methods. In Sec. \ref{a:unbalanced} we present results with ResNet20 models trained on disjoint subsets of CIFAR100, where CCA Merge also achieves smaller drops in accuracy than the other methods.
%Models merged with CCA Merge have significantly higher accuracies than models merged with any othe method, and this is true across all models and widths considered. Differences in accuracy up to 25\% (VGG11$\times 1$ width) can be observed between CCA Merge and the second-best performing method, with CCA Merge outperforming the other baselines across all settings. 

For both VGG and ResNet architectures as well as for all considered datasets the added flexibility of CCA Merge over permutation-based methods seems to benefit the merged models. Aligning models using linear combinations allows CCA Merge to better model relationships between neurons and to take into account features that are distributed across multiple neurons. In addition to the raw performance benefits, CCA Merge seems to be more stable across different model widths as well as across different random initializations and data order and augmentation.

% \begin{table}[h]
%     % \vspace{-10pt}
%     \centering
%     \begin{tabular}{l|l l}
%         Method & Top 1 Acc. (\%) & Top 5 Acc. (\%)  \\
%         \noalign{\smallskip}
%         \hline
%         \noalign{\smallskip}
%         \textcolor{gray}{Base models avg.} & \textcolor{gray}{82.09 \small{$\pm$0.13}} & \textcolor{gray}{95.21 \small{$\pm$0.09}}\\
%         \textcolor{gray}{Ensemble}         & \textcolor{gray}{83.51 \small{$\pm$0.01}} & \textcolor{gray}{95.9 \small{$\pm$0.03}}\\
%         \noalign{\smallskip}
%         \hline
%         \noalign{\smallskip}
%         Direct averaging & 2.94 \small{$\pm$0.12} & 10.52 \small{$\pm$0.60}\\
%         Permute          & 71.84 \small{$\pm$0.53} & 91.40 \small{$\pm$0.24}\\
%         OT Fusion          & 71.64 \small{$\pm$0.73} & 91.15 \small{$\pm$0.24}\\
%         Matching Weights & 69.37 \small{$\pm$0.48} & 90.53 \small{$\pm$0.24}\\
%         \rowcolor{gray!25}\textbf{ CCA Merge (ours)} & \textbf{76.38 \small{$\pm$0.20}} & \textbf{93.03 \small{$\pm$0.21}}\\
%     \end{tabular}
%     \caption{ResNet18x4 trained on ImageNet200 - Accuracies and standard deviations from 3 different merges of 2 models are presented. Even on this significantly harder image classification task CCA Merge outperforms past model merging methods for both top 1 and top 5 accuracies. In the case of top 5 accuracy CCA Merge remarkably approaches the performance of model ensembles.}
%     \vspace{-5pt}
%     \label{t:results_imagenet}
%     \vspace{-5pt}
% \end{table}

\begin{table}[ht]
    % \vspace{-10pt}
    \centering
    \begin{tabular}{l|l l}
        Method & Top 1 Acc. (\%) & Top 5 Acc. (\%)  \\
        \noalign{\smallskip}
        \hline
        \noalign{\smallskip}
        \textcolor{gray}{Base models avg.} & \textcolor{gray}{75.44 \small{$\pm$0.06}} & \textcolor{gray}{92.17 \small{$\pm$0.05}}\\
        \textcolor{gray}{Ensemble}         & \textcolor{gray}{77.62 \small{$\pm$0.07}} & \textcolor{gray}{93.48 \small{$\pm$0.01}}\\
        \noalign{\smallskip}
        \hline
        \noalign{\smallskip}
        Direct averaging & 0.12 \small{$\pm$0.03} & 0.64 \small{$\pm$0.01}\\
        Permute          & 51.45 \small{$\pm$1.02} & 76.96 \small{$\pm$0.76}\\
        OT Fusion          & 50.55 \small{$\pm$0.98} & 76.35 \small{$\pm$0.97}\\
        Matching Weights & 45.41 \small{$\pm$0.33} & 72.78 \small{$\pm$0.42}\\
        \rowcolor{gray!25}\textbf{ CCA Merge (ours)} & \textbf{63.61 \small{$\pm$0.22}} & \textbf{85.41 \small{$\pm$0.25}}\\
    \end{tabular}
    \caption{ResNet18x4 trained on ImageNet - Accuracies and standard deviations from 3 different merges of 2 models are presented. Even on this significantly harder image classification task CCA Merge outperforms past model merging methods for both top 1 and top 5 accuracies, helping to reduce the gap between model merging methods and model ensembles.
    %In the case of top 5 accuracy CCA Merge remarkably approaches the performance of model ensembles.
    }
    \vspace{-5pt}
    \label{t:results_imagenet}
    \vspace{-5pt}
\end{table}

\subsection{CCA merging finds better common representations between many models}\label{ss:many_models}
% [PLOTS OF BARRIER AS NUM MODELS INCREASE]
In this section, we present our results related to the merging of many models, a significantly harder task. This constitutes the natural progression to the problem of merging pairs of models and is a more realistic setting for distributed or federated learning applications where there are often more than 2 models. Furthermore, aligning populations of neural networks brings us one step closer to finding the common learned features that allow different neural networks to perform equally as well on complex tasks despite having different initializations, data orders, and data augmentations.

As previously mentioned, the problem of merging many models is often ignored by past works except for the settings in which linear mode connectivity is easily obtained. \citet{ainsworth2023_git-rebasin} introduced ``Merge Many'', an adaptation of Matching Weights for merging a set of models. In Merge Many each model in the set is sequentially matched to the average of all the others until convergence. A simpler way of extending any model merging method to the many models setting is to choose one of the models in the group as the \emph{reference model} and to align every other network in the group to it. Then the reference model and all the other aligned models can be merged. It is by using this ``all-to-one'' merging that we extend CCA Merge, Permute, OT Fusion and Matching Weights to the many model settings. ZipIt! is naturally able to merge multiple models since it aggregates all neurons and merges them until the desired size is obtained.

\begin{figure}[ht]
\begin{center}
\includegraphics[width=0.475\textwidth]{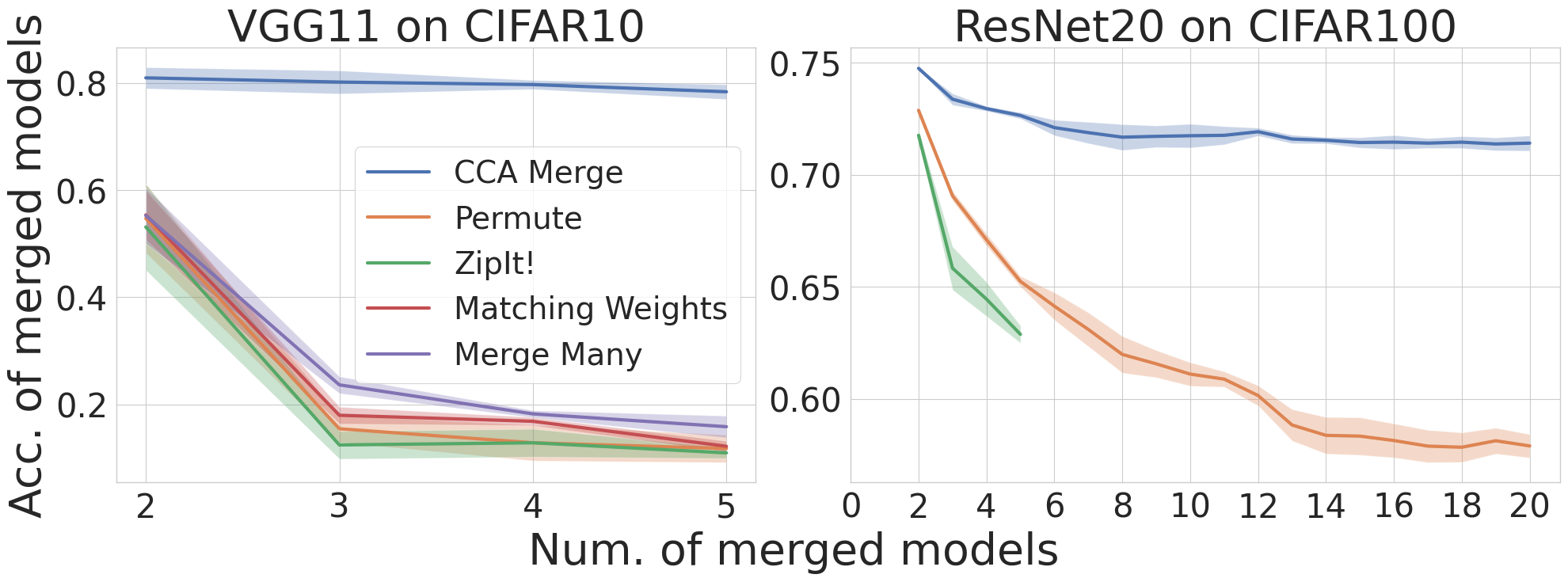}
\end{center}
\vspace{-12pt}
\caption{Accuracies of averaging multiple models after feature alignment with different merging methods. Mean and standard deviation across 4 random seeds are shown.}
\vspace{-10pt}
\label{fig:multiple_models}
\end{figure}

In Fig. \ref{fig:multiple_models} we show the accuracies of the merged models as the number of models being merged increases. For both VGG and ResNet architectures aligning model weights with CCA continues to yield better performing merged models. In fact, models merged with CCA Merge applied in an all-to-one fashion maintain their accuracy relatively well while the ones merged with other methods see their accuracies drop significantly. In the VGG case, the drop for other methods is drastic, all merged models having less than 20\% accuracy when more than 3 models are being merged which constitutes a decrease of more than 25\% from the 2 models merged scenario. CCA Merge on the other hand suffers a drop in accuracy of less than 3\% when going from merging 2 models to 5, staying around the 80\% mark.
We also note that despite being designed specifically for the many models setting, Merge Many performs only slightly better than its 2 model counterpart (Matching Weights applied in an all-to-one fashion).

\begin{table*}[ht]
    \centering
    \begin{tabular}{l|l|l|l}
        Method & (1) 80\%-20\% & (2) Dirichlet & (3) 50 classes\\
        \noalign{\smallskip}
        \hline
        \noalign{\smallskip}
        \textcolor{gray}{Base models avg.}  & \textcolor{gray}{65.66 \small{$\pm$0.71\%}} & \textcolor{gray}{59.98 \small{$\pm$1.80\%}} & \textcolor{gray}{41.42 \small{$\pm$0.54\%}} \\
        \textcolor{gray}{Ensemble}          & \textcolor{gray}{77.84 \small{$\pm$0.23\%}} & \textcolor{gray}{73.77 \small{$\pm$0.44\%}}  & \textcolor{gray}{69.91 \small{$\pm$0.49\%}} \\
        \noalign{\smallskip}
        \hline
        \noalign{\smallskip}
        Direct averaging  & 11.40 \small{$\pm$1.62\%} & 20.55 \small{$\pm$3.07\%}  & 16.90 \small{$\pm$2.02\%} \\
        Permute           & 62.11 \small{$\pm$0.30\%} & 58.45 \small{$\pm$1.76\%}  & 43.82 \small{$\pm$1.31\%} \\
        OT Fusion         & 61.56 \small{$\pm$0.21\%} & 57.67 \small{$\pm$1.49\%}  & 43.02 \small{$\pm$1.27\%} \\
        Matching Weights   & 58.18 \small{$\pm$0.68\%} & 55.87 \small{$\pm$1.80\%}  & 41.15 \small{$\pm$1.45\%} \\
        ZipIt!            & 61.41 \small{$\pm$0.51\%} & 57.97 \small{$\pm$1.29\%}  & \textbf{55.08 \small{$\pm$0.70\%}} \\
        \rowcolor{gray!25}\textbf{CCA Merge (ours)}  & \textbf{66.35 \small{$\pm$0.19\%}} & \textbf{60.38 \small{$\pm$1.68\%}} & 46.57 \small{$\pm$0.76\%} \\
    \end{tabular}
\vspace{-5pt}
\caption{ResNet20$\times 8$ trained on 3 different splits of CIFAR100 - Accuracies and standard deviations from 4 different merges of 2 models are presented. When the models being merged have learned different features from disjoint sets of the training data but with all the classes (splits 1 and 2) CCA Merge is the only model merging method which outperforms the average of the base models. In the case where the models being merged were trained on disjoint subsets of the classes (split 3) CCA Merge still outperforms past model merging methods except for ZipIt!. However ZipIt! allows same-model neuron merging, making a direct comparison with the other methods, including ours, somewhat unfair.}
\label{t:split_data_results}
\vspace{-8pt}
\end{table*}

For ResNets, the accuracy of models merged with Permute drops by \texttildee15\% when going from merging 2 models to 20. While less drastic than in the VGG case this decrease in performance is still significant. ZipIt! displays a slightly more pronounced drop when going from 2 models merged to 5. CCA Merge on the other hand is a lot more robust, incurring a less than 4\% drop in accuracy even as the number of merged models is increased to 20. Additionally, accuracy values for models merged by CCA Merge seem to plateau sooner than those for Permute.

These results suggest that CCA Merge is significantly better than past methods at finding the ``common features'' learned by groups of neural networks and aligning them. The limitations of permutation-based methods in taking into account complex relationships between neurons from different models are highlighted in this context. Here it is harder to align features given that there are more of them to consider and therefore easier to destroy the features when averaging them.

\paragraph{Permutations and feature mismatches}
To explain the success of CCA Merge in the multi-model setting, we have examined feature mismatches between networks. When merging models $\mathcal{A}$, $\mathcal{B}$ and $\mathcal{C}$, we choose one reference model, say model $\mathcal{A}$, align $\mathcal{B}$ and $\mathcal{C}$ to $\mathcal{A}$ and then merge by averaging. This results in an indirect matching of models $\mathcal{B}$ and $\mathcal{C}$ through the reference model $\mathcal{A}$. Neurons $i$ from $\mathcal{B}$ and $j$ from $\mathcal{C}$ are matched indirectly through $\mathcal{A}$ if they both are matched to the same neuron $k$ from $\mathcal{A}$. Ideally, this indirect matching should be the same as the direct matching resulting from aligning $\mathcal{B}$ and $\mathcal{C}$ directly. This would mean that the same features from $\mathcal{B}$ and $\mathcal{C}$ are matched regardless of whether they are merged together or with a third model.
However, this does not seem to be the case for permutation-based methods: over 50\% of the neurons from $\mathcal{B}$ don't get matched to the same neuron from $\mathcal{C}$ when aligned directly or indirectly through $\mathcal{A}$. Additionally, the Frobenius norm of the differences between the direct and indirect matching matrices is significantly lower for CCA Merge than for a permutation-based method. This suggests CCA Merge generates fewer feature mismatches in the multi-model setting, explaining in part its success over permutation-based methods. For detailed results and further information see App. \ref{a:multi_model}.

\subsection{CCA Merge is better at combining learned features from different data splits}\label{ss:split_data}
\vspace{-5pt}
In this section we consider the more realistic setting where the models are trained on disjoint splits of the data, therefore they're expected to learn (at least some) different features. Such a set-up is natural in the context of federated or distributed learning. We consider ResNet20 models trained on 3 different data splits of the CIFAR100 training dataset. The first (1) data split is the one considered in \citet{ainsworth2023_git-rebasin, jordan2023repair} where one model is trained on 80\% of the data from the first 50 classes of the CIFAR100 dataset and 20\% of the data from the last 50 classes, the second model being trained on the remaining examples. In the second (2) data split we use samples from a Dirichlet distribution with parameter vector $\alpha = (0.5, 0.5)$ to subsamble each class in the dataset and create 2 disjoint data splits, one for each model to be trained on. Lastly, with the third (3) data split we consider the more extreme scenario from \citet{stoica2024zipit} where one model is trained on 100\% of the data from 50 classes, picked at random, and the second one is trained on the remaining classes, with no overlap. For this last setting, in order for both models to have a common output space they were trained using the CLIP \cite{radford2021_clip} embeddings of the class names as training objectives. In Table \ref{t:split_data_results} we report mean and standard deviation of accuracies across 4 different model pairs.

For the first two data splits CCA Merge outperforms the other methods, beating the second best method by \texttildee4\% and \texttildee2\% on the first and second data splits respectively. For the third data split CCA Merge is the second best performing method after ZipIt!. However, ZipIt! was designed for this specific setting and, as we previously noted, it allows the merging of features from the same network to reduce redundancies, thus making it more flexible than the other methods which only perform ``alignment''. In all cases CCA Merge outperforms or is comparable with the base models average indicating that, to some extent, our method successfully combines different learned features from the two models.

% \subsection{CCA Merge offers better starting points for fine-tuning}
% Lastly we also consider the case where we continue training after merging the models. 
\vspace{-7pt}
\section{Discussion and Conclusion}
\vspace{-5pt}
Recent model fusion successes exploit inter-model relationships between neurons by modelling them as permutations before fusing models.
Here, we argue that, while assuming a one-to-one correspondence between neurons yields interesting merging methods, it is rather limited as not all neurons from one network have an exact match with a neuron from another network.
Our proposed \emph{CCA Merge} takes the approach of linearly transforming model parameters beyond permutation-based optimization. This added flexibility allows our method to
outperform recent competitive baselines when merging pairs of models trained on the same data or on disjoint splits of the data (Tables \ref{t:results_same_data}, \ref{t:results_imagenet} and \ref{t:split_data_results}). Furthermore, when considering the harder task of merging many models, CCA Merge models showcase remarkable accuracy stability as the number of models merged increases, while past methods suffer debilitating accuracy drops. This suggests a path towards achieving \emph{strong linear connectivity} between a set of models, which is hard to do with permutations \citep{sharma2024simultaneous}.
One limitation of our method is that it requires input data to align the models. The forward pass to compute the activations increases computational costs and in some settings such a ``shared'' dataset might not be available. We discuss this further in App. \ref{a:comp_costs}.

Merging many models successfully, without incurring an accuracy drop, is one of the big challenges in this area of research. Our method, CCA Merge, makes a step in the direction of overcoming this challenge. As future work, it would be interesting to further study the common representations learned by populations of neural networks. Also, an interesting future research avenue is to test CCA Merge with models trained on entirely different datasets, to test its limits in terms of combining different features.

\section*{Acknowledgements}
This work was partially funded by the Natural Sciences and Engineering Research Council of Canada (NSERC) CGS D 569345 - 2022 scholarship [S.H.]; FRQNT-NSERC grant 2023-NOVA-329125 [E.B. \& G.W.]; Canada CIFAR AI Chair, NSF DMS grant 2327211 and NSERC Discovery grant 03267 [G.W.]. 
This work is also supported by resources from Compute Canada and Calcul Québec. 
The content is solely the responsibility of the authors and does not necessarily represent the views of the funding agencies.
\section*{Impact Statement}
This paper presents work whose goal is to advance the field of Machine Learning, specifically the sub-field interested in merging models which were trained from scratch, potentially from different initializations, on different data splits or datasets or following different training procedures.

As deep learning has gained in popularity, many open-source models have become available online and are increasingly being used in research and in industry. In many cases these models do not come from the same initialization and the training data and pipelines are varied and in some cases unavailable. 
Advances in model merging might provide more storage and compute efficient ways of taking advantage of such open-sourced models, thus reducing resource consumption. Furthermore, model merging methods have the potential of allowing us to find the fundamental shared features learned by different models, thus unifying different data representations and furthering our understanding of deep learning models.

Any other societal and ethical impacts of our work are likely to be the same as for any work advancing the field of Machine Learning in general.

\bibliography{main}
\bibliographystyle{icml2024}

%%%%%%%%%%%%%%%%%%%%%%%%%%%%%%%%%%%%%%%%%%%%%%%%%%%%%%%%%%%%%%%%%%%%%%%%%%%%%%%
%%%%%%%%%%%%%%%%%%%%%%%%%%%%%%%%%%%%%%%%%%%%%%%%%%%%%%%%%%%%%%%%%%%%%%%%%%%%%%%
% APPENDIX
%%%%%%%%%%%%%%%%%%%%%%%%%%%%%%%%%%%%%%%%%%%%%%%%%%%%%%%%%%%%%%%%%%%%%%%%%%%%%%%
%%%%%%%%%%%%%%%%%%%%%%%%%%%%%%%%%%%%%%%%%%%%%%%%%%%%%%%%%%%%%%%%%%%%%%%%%%%%%%%
\newpage
\appendix
\onecolumn
% \section{Appendix}
\section{Additional Practical Considerations for CCA Merge}\label{a:pract_cons}
\subsection{Aligning $\mathcal{B}$ to $\mathcal{A}$ rather than transforming both}\label{a:align_ba}
Given that CCA naturally finds a common representation space which maximally correlates the activations of both models being merged, $\mathbf{X}^\mathcal{A}$ and $\mathbf{X}^\mathcal{B}$, it is natural to consider merging the models in this common space. This would be done by transforming the parameters of model $\mathcal{A}$ with the found transformations $\mathbf{P}_i^\mathcal{A}$ and transforming the parameters of model $\mathcal{B}$ with transformations $\mathbf{P}_i^\mathcal{B}$ and then averaging both transformed models. However, the reason for applying transformation $\mathbf{T}=\left(\mathbf{P}^\mathcal{B}\cdot\mathbf{P}^\mathcal{A}\right)^\top$ to model $\mathcal{B}$ to align it to $\mathcal{A}$ and merging both models in the representation space of model $\mathcal{A}$ instead of transforming both the weights of model A and model B and merging in the common space found by CCA is because of the ReLU non-linearity. The common representation space found by CCA has no notion of directionality, and might contain important information even in negative orthants (high-dimensional quadrants where at least one of the variables has negative values) that might get squashed to zero by ReLU non-linearities. The representation space of model $\mathcal{A}$ doesn’t have this problem.

\subsection{Regularized CCA}\label{a:reg_cca}
The closed form CCA solution requires inverting the scatter matrices $\mathbf{S}^{\mathcal{A}\mathcal{A}}$ and $\mathbf{S}^{\mathcal{B}\mathcal{B}}$ which can lead to poor performance and instability when the eigenvalues of these matrices are small. To make CCA more robust, the identity matrix $\mathbf{I}$ scaled by a regularization hyper-parameter $\gamma$ is added to the two scatter matrices. Therefore to complete the CCA computation the matrices $\mathbf{S}^{\mathcal{A}\mathcal{A}} + \gamma\mathbf{I}$ and $\mathbf{S}^{\mathcal{B}\mathcal{B}} + \gamma\mathbf{I}$ are used instead of $\mathbf{S}^{\mathcal{A}\mathcal{A}}$ and $\mathbf{S}^{\mathcal{B}\mathcal{B}}$.

To chose the hyper-parameter $\gamma$ for any experiment, i.e. a combination of model architecture and data set / data split, we train additional models from scratch and conduct the merging with different $\gamma$ values. The $\gamma$ value leading to the best merged accuracy is kept and applied to the other experiments with CCA Merge. The models used to select $\gamma$ are discarded to avoid over-fitting and the CCA Merge accuracies are reported for new sets of models.

\section{Further empirical analysis of matching matrices}\label{a:non_opt_matches_permute}
\subsection{Non-optimal matches of permutation-based methods}
When merging 2 models, $\mathcal{A}$ and $\mathcal{B}$, with Permute, a large percentage (25-50\%) of the neurons from model $\mathcal{A}$ do not get matched with their highest correlated neuron from model $\mathcal{B}$ by the permutation matrix, we call these non-optimal matches. In other words, for neurons in $\mathcal{A}$ that have a non-optimal match, a higher correlated neuron from $\mathcal{B}$ exists to which that $\mathcal{A}$ neuron isn't matched. Since permutation-based methods aim to optimize an overall cost, the found solutions match some neurons non-optimally in order to obtain a better overall score. However, since permutation-based methods either completely align two features (value of 1 in the matching matrix) or not at all (value of 0), the merging completely ignores the relationship between these highly correlated but not matched neurons.

\begin{figure}[ht!]
\begin{center}
\includegraphics[width=0.75\textwidth]{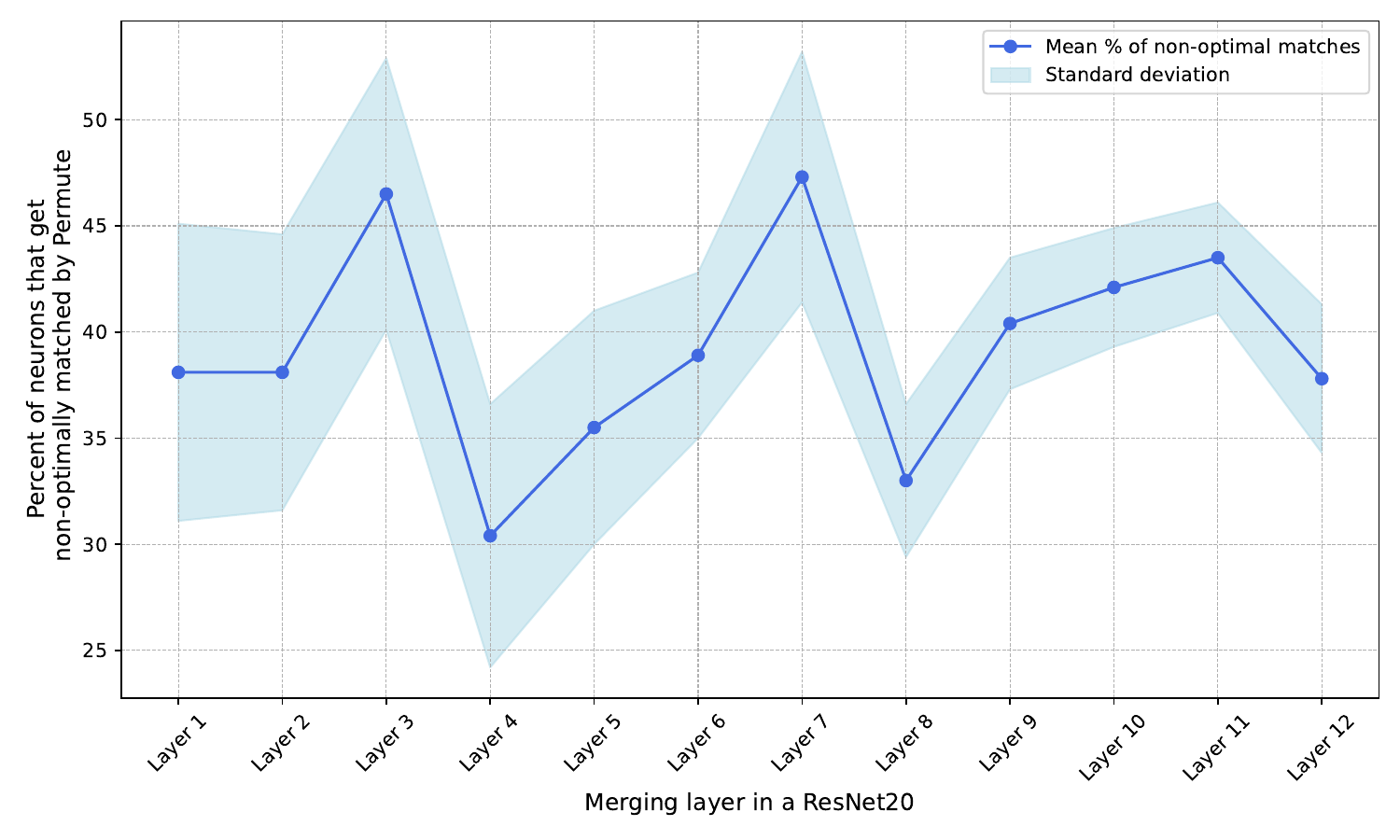}
\end{center}
% \vspace{-12pt}
\caption{Percent (\%) of non-optimal matches when merging ResNet20x8 models trained on CIFAR100. The mean and standard deviation across 15 possible 2-model merges out of a group of 6 models fully trained from different initializations are shown.}
% \vspace{-5pt}
\label{fig:non_opt_matches_permute}
\end{figure}

In Fig. \ref{fig:non_opt_matches_permute} we present the percent of neurons from net $\mathcal{A}$ that get non-optimally matched to a neuron from net $\mathcal{B}$ for all merging layers inside a ResNet20 trained on CIFAR100. We can see from these results that the limiting nature of permutations causes Permute to disregard some meaningful relationships between learned features, hindering the merged model accuracy.

\subsection{Highest correlated neurons are associated to top CCA Merge transformation coefficients}

\begin{figure}[ht]
\begin{center}
\includegraphics[width=0.75\textwidth]{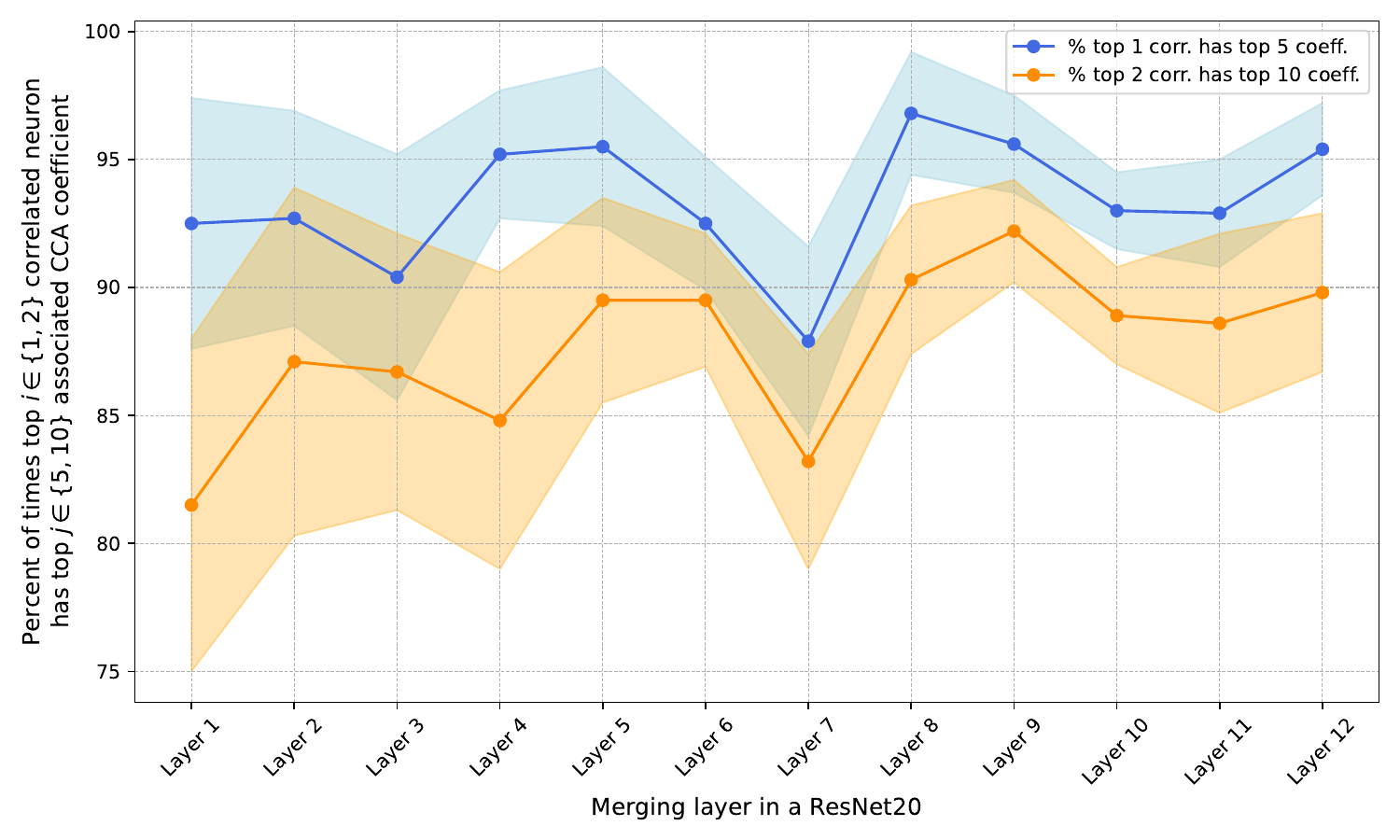}
\end{center}
% \vspace{-12pt}
\caption{Percent (\%) of top 1 and top 2 correlated neurons that have top 5 and top 10 CCA Merge coefficients respectively when merging ResNet20x8 models trained on CIFAR100. The mean and standard deviation across 15 possible 2-model merges out of a group of 6 models fully trained from different initializations are shown.}
% \vspace{-5pt}
\label{fig:top_cca_coeffs}
\end{figure}

Since CCA Merge combines linear combinations of neurons instead of just individual ones, we can't run the exact same analysis as we did in the previous section for Permute. However, we have looked at whether or not, for each neuron from model $\mathcal{A}$, the top 1 and top 2 correlated neuron from model $\mathcal{B}$ (i.e. neurons with the highest or second highest correlations respectively) get assigned high coefficients in the CCA Merge transformation matrices. We report these values in the Fig. \ref{fig:top_cca_coeffs}.

In the vast majority of cases the highest correlated neuron from model $\mathcal{B}$ gets assigned one of the 5 highest coefficients in the CCA Merge transform, and the top 2 correlated neuron gets assigned one of the 10 highest coefficients of the transform. These results showcase how CCA Merge better accounts for the relationships between neurons of different models during the merging procedure.

\section{Additional Results}\label{a:same_data}
Here we present the extended Table \ref{t:results_same_data} results, where we also include results for widths $\times 2$ and $\times 4$ for both VGG and ResNet models. The same overall conclusions hold with CCA Merge performing generally better than all other considered baselines. CCA Merge's standard deviation isn't always the lowest however as width increases but it's comparable with the other methods. As noted in the main text we ran into out-of-memory issues when running ZipIt! for VGG models with width greater than $\times 2$.

\begin{table}[!ht]
\caption{VGG11/CIFAR10 - Accuracies and standard deviations after merging 2 models}
\vspace{-10pt}
\label{t:vgg11_cifar10}
\begin{center}
\begin{tabular}{l|l l l l}
            Width multiplier & $\times 1$ & $\times 2$ & $\times 4$ & $\times 8$\\
            \noalign{\smallskip}
            \hline
            \noalign{\smallskip}
            \textcolor{gray}{Base models avg.} & \textcolor{gray}{87.27 \small{$\pm$0.25\%}} & \textcolor{gray}{87.42 \small{$\pm$0.86\%}} & \textcolor{gray}{87.84 \small{$\pm$0.21\%}} & \textcolor{gray}{88.20 \small{$\pm$0.45\%}}\\
            \textcolor{gray}{Ensemble}         & \textcolor{gray}{89.65 \small{$\pm$0.13\%}} & \textcolor{gray}{89.74 \small{$\pm$0.44\%}} & \textcolor{gray}{90.12 \small{$\pm$0.16\%}} & \textcolor{gray}{90.21 \small{$\pm$0.24\%}}\\
            \noalign{\smallskip}
            \hline
            \noalign{\smallskip}
            Direct averaging & 10.54 \small{$\pm$0.93\%} & 10.28 \small{$\pm$0.48\%} & 10.00 \small{$\pm$0.01\%} & 10.45 \small{$\pm$0.74\%}\\
            Permute          & 54.39 \small{$\pm$6.45\%} & 63.32 \small{$\pm$1.12\%} & 64.81 \small{$\pm$1.99\%} & 62.58 \small{$\pm$3.31\%}\\
            OT Fusion        & 53.86 \small{$\pm$10.4\%} & 65.97 \small{$\pm$2.13\%} & 66.34 \small{$\pm$3.17\%} & 68.32 \small{$\pm$3.13\%}\\ 
            Matching Weights & 55.40 \small{$\pm$4.67\%} & 66.98 \small{$\pm$1.96\%} & 71.92 \small{$\pm$2.21\%} & 73.74 \small{$\pm$1.77\%}\\
            ZipIt!           & 52.93 \small{$\pm$6.37\%} & 60.73 \small{$\pm$2.07\%} & - & - \\
            \rowcolor{gray!25}\textbf{CCA Merge (ours)}        & \textbf{82.65 \small{$\pm$0.73\%}} & \textbf{83.31 \small{$\pm$1.05\%}} & \textbf{85.54 \small{$\pm$0.51\%}} & \textbf{84.36 \small{$\pm$2.09\%}} \\
        \end{tabular}
\end{center}
\end{table}

\begin{table}[!ht]
\caption{ResNet20/CIFAR100 - Accuracies and standard deviations after merging 2 models}
\vspace{-10pt}
\label{t:resnet20_cifar100}
\begin{center}
\begin{tabular}{l|l l l l}
            Width multiplier & $\times 1$ & $\times 2$ & $\times 4$ & $\times 8$\\
            \noalign{\smallskip}
            \hline
            \noalign{\smallskip}
            \textcolor{gray}{Base models avg.} & \textcolor{gray}{69.21 \small{$\pm$0.22\%}} & \textcolor{gray}{74.22 \small{$\pm$0.14\%}} & \textcolor{gray}{77.28 \small{$\pm$0.34\%}} & \textcolor{gray}{78.77 \small{$\pm$0.28\%}}\\
            \textcolor{gray}{Ensemble}         & \textcolor{gray}{73.51 \small{$\pm$0.20\%}} & \textcolor{gray}{77.57 \small{$\pm$0.19\%}} & \textcolor{gray}{79.90 \small{$\pm$0.08\%}} & \textcolor{gray}{80.98 \small{$\pm$0.21\%}}\\
            \noalign{\smallskip}
            \hline
            \noalign{\smallskip}
            Direct averaging & 1.61 \small{$\pm$0.16\%}  & 2.67 \small{$\pm$0.16\%} & 5.13 \small{$\pm$0.52\%} & 14.00 \small{$\pm$1.66\%}\\
            Permute          & 28.76 \small{$\pm$2.20\%} & 49.45 \small{$\pm$0.41\%} & 64.65 \small{$\pm$0.34\%} & 72.90 \small{$\pm$0.08\%}\\
            OT Fusion        & 29.05 \small{$\pm$2.55\%} & 48.74 \small{$\pm$1.11\%} & 64.07 \small{$\pm$0.38\%} & 72.45 \small{$\pm$0.08\%}\\
            Matching Weights & 21.38 \small{$\pm$4.36\%} & 44.85 \small{$\pm$0.66\%} & 64.66 \small{$\pm$0.45\%} & 74.29 \small{$\pm$0.51\%}\\
            ZipIt!           & 25.26 \small{$\pm$2.30\%} & 47.72 \small{$\pm$0.53\%} & 63.69 \small{$\pm$0.31\%} & 72.47 \small{$\pm$0.41\%}\\
            % \hline
            \rowcolor{gray!25}\textbf{CCA Merge (ours)}        & \textbf{31.79 \small{$\pm$1.97\%}} & \textbf{54.26 \small{$\pm$1.00\%}} & \textbf{68.75 \small{$\pm$0.22\%}} & \textbf{75.06 \small{$\pm$0.18\%}} \\
        \end{tabular}
\end{center}
\vspace{-5pt}
\end{table}

\section{VGG Results with REPAIR}
In this section, specifically in Table \ref{t:vgg11_cifar10_repair}, we present the results for merging VGG networks with REPAIR \cite{jordan2023repair} applied to mitigate the variance collapse phenomenon. Since the standard VGG architecture doesn't contain normalization layers it isn't as straightforward as just resetting the BatchNorm statistics as it was for ResNets.
Applying REPAIR seems to greatly help past methods and all methods are now comparable in terms of accuracy, with Matching Weights \cite{ainsworth2023_git-rebasin} being slightly better, but the standard deviations overlapping with CCA Merge.
The great performance of CCA Merge without REPAIR suggests that perhaps models merged with our method do not suffer from variance collapse to the same extent as models merged with permutation-based methods.
One interesting thing to note is that REPAIR also significantly helps Direct Averaging, where the networks aren't aligned before merging. Without REPAIR, Direct Averaging performed no better than random, however with REPAIR the $\times8$ models achieve $>70\%$ accuracy even without any sort of alignment.

Since CCA Merge uses general invertible linear transformations for alignment instead of permutations, the model being aligned can have its functionality altered post alignment. Therefore, when applying REPAIR to models merged with CCA Merge, we do not set the mean and standard deviation of the averaged model's activations to be the average of the mean and standard deviation of the models being merged (standard REPAIR). Instead, we simply reset the mean and standard deviation of the neurons to be the same as the ones of the reference model, i.e. the one to which no transformation is applied and therefore its functionality wasn't altered. To make sure that the comparison to the other methods is fair we have also tried doing this for all the other methods, with the results being around 1\% worse than simply applying standard REPAIR (results show in Table \ref{t:vgg11_cifar10_repair}).

\begin{table}[!ht]
\caption{VGG11/CIFAR10 - Accuracies and standard deviations after merging 2 models with REPAIR}
\vspace{-10pt}
\label{t:vgg11_cifar10_repair}
\begin{center}
\begin{tabular}{l|l l l l}
            Width multiplier & $\times 1$ & $\times 2$ & $\times 4$ & $\times 8$\\
            \noalign{\smallskip}
            \hline
            \noalign{\smallskip}
            \textcolor{gray}{Base models avg.} & \textcolor{gray}{87.27 \small{$\pm$0.25\%}} & \textcolor{gray}{87.42 \small{$\pm$0.86\%}} & \textcolor{gray}{87.84 \small{$\pm$0.21\%}} & \textcolor{gray}{88.20 \small{$\pm$0.45\%}}\\
            \textcolor{gray}{Ensemble}         & \textcolor{gray}{89.65 \small{$\pm$0.13\%}} & \textcolor{gray}{89.74 \small{$\pm$0.44\%}} & \textcolor{gray}{90.12 \small{$\pm$0.16\%}} & \textcolor{gray}{90.21 \small{$\pm$0.24\%}}\\
            \noalign{\smallskip}
            \hline
            \noalign{\smallskip}
            Direct averaging & 29.89 \small{$\pm$0.58\%} & 43.16 \small{$\pm$5.19\%} & 56.83 \small{$\pm$4.34\%} & 73.75 \small{$\pm$3.12\%}\\
            Permute          & 84.56 \small{$\pm$0.30\%} & 85.72 \small{$\pm$0.87\%} & 87.57 \small{$\pm$0.13\%} & 88.35 \small{$\pm$0.62\%}\\
            OT Fusion        & 83.33 \small{$\pm$0.32\%} & 84.18 \small{$\pm$1.12\%} & 86.34 \small{$\pm$0.38\%} & 87.09 \small{$\pm$0.30\%}\\ 
            Matching Weights & \textbf{85.62 \small{$\pm$0.38\%}} & \textbf{86.98 \small{$\pm$0.34\%}} & \textbf{88.68 \small{$\pm$0.18\%}} & \textbf{88.94 \small{$\pm$0.14\%}}\\
            \rowcolor{gray!25}\textbf{CCA Merge (ours)}        & \textbf{85.02 \small{$\pm$0.27\%}} & \textbf{86.51 \small{$\pm$0.37\%}} & 87.38 \small{$\pm$0.55\%} & \textbf{88.17 \small{$\pm$0.67\%}} \\
        \end{tabular}
\end{center}
\end{table}

\section{Why does CCA Merge outperform permutation-based methods in the multi-model setting?}\label{a:multi_model}
\subsection{When merging multiple models, feature mismatches of permutation-based methods get compounded}
When merging more than two models, say $\mathcal{A}$, $\mathcal{B}$ and $\mathcal{C}$ in the 3-model case, we choose one reference model, WLOG model $\mathcal{A}$, align all other models to that one and then merge by averaging (all-to-one merging). Let $\mathbf{T}_{\mathcal{BA}}$ denote the transformation aligning a given layer of model $\mathcal{B}$ to the same layer of model $\mathcal{A}$ and $\mathbf{T}_{\mathcal{CA}}$ the transformation aligning $\mathcal{C}$ to $\mathcal{A}$ at that same layer, we can also align $\mathcal{A}$ to $\mathcal{B}$ (resp. $\mathcal{C}$) by taking the inverse of $\mathbf{T}_{\mathcal{BA}}$ (resp. $\mathbf{T}_{\mathcal{CA}}$) which we denote $\mathbf{T}_{\mathcal{AB}}=\mathbf{T}_{\mathcal{BA}}^{-1}$ (resp. $\mathbf{T}_{\mathcal{AC}}=\mathbf{T}_{\mathcal{CA}}^{-1}$). This alignment to $\mathcal{A}$ also gives rise to an indirect matching between models $\mathcal{B}$ and $\mathcal{C}$ through $\mathcal{A}$, since we can align $\mathcal{C}$ to $\mathcal{A}$ with $\mathbf{T}_{\mathcal{CA}}$ and then apply the transformation $\mathbf{T}_{\mathcal{AB}}$ to align it to $\mathcal{B}$, we use $\mathbf{T}_{\mathcal{CAB}}$ to denote the transformation of this indirect alignment. In other words, neurons $i$ from $\mathcal{B}$ and $j$ from $\mathcal{C}$ are matched indirectly through $\mathcal{A}$ if they both are matched to the same neuron $k$ from $\mathcal{A}$. To see why multi-model merging fails for permutations we can look at how this indirect matching of $\mathcal{B}$ and $\mathcal{C}$ ($\mathbf{T}_{\mathcal{CAB}}$) compares to the direct matching of $\mathcal{B}$ and $\mathcal{C}$ found by directly optimizing for the transformation matrix $\mathbf{T}_{\mathcal{CB}}$. It turns out that for permutation-based methods these matrices differ significantly, in fact the majority of neurons from net $\mathcal{C}$, between 50-80\% on average, do not get matched with the same neuron from $\mathcal{B}$ if we use $\mathbf{T}_{\mathcal{CAB}}$ versus if we use the direct matching $\mathbf{T}_{\mathcal{CB}}$.

\begin{figure}[ht]
\begin{center}
\includegraphics[width=0.75\textwidth]{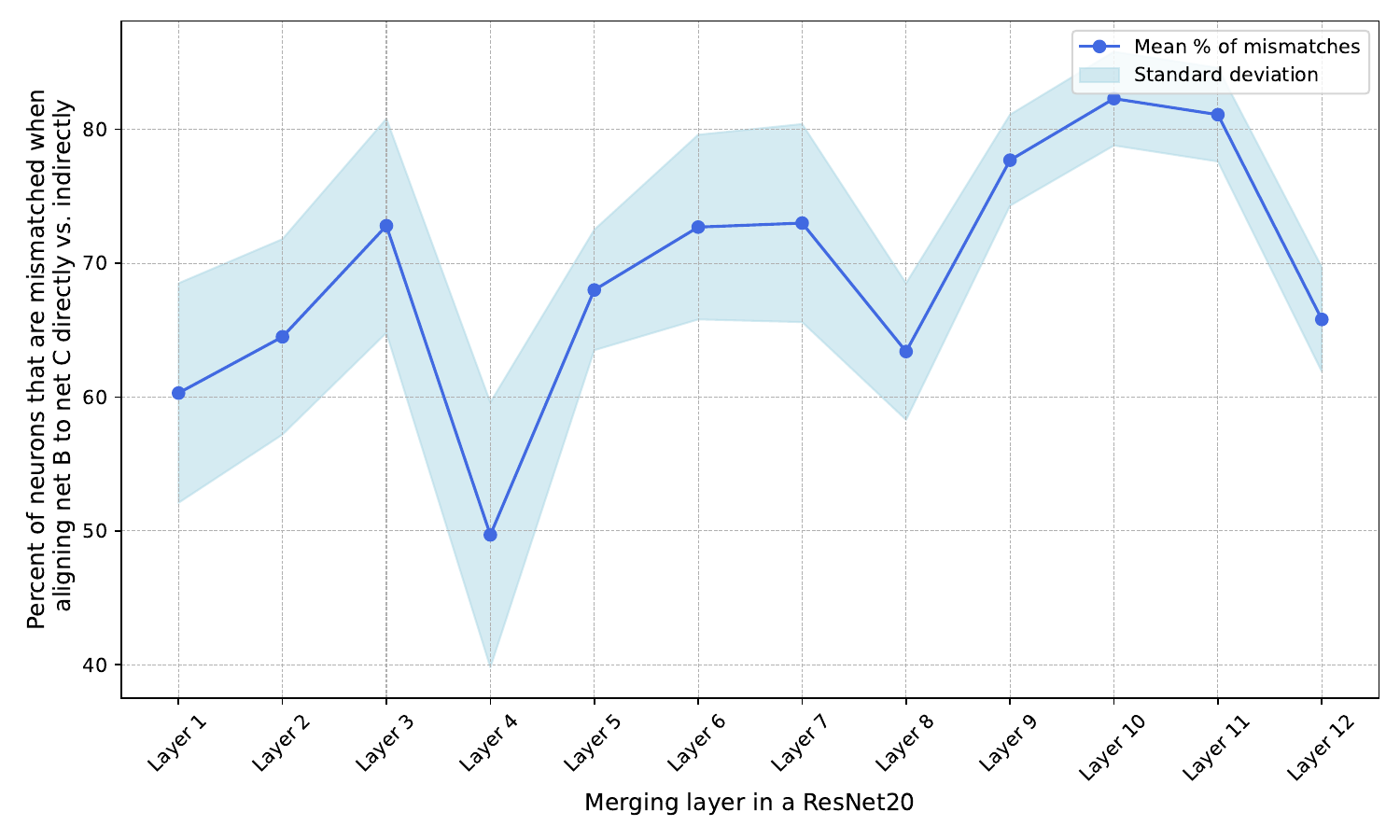}
\end{center}
% \vspace{-12pt}
\caption{Percent (\%) of neurons getting mismatched when merging ResNet20x8 models trained on CIFAR100. The mean and standard deviation across 20 possible 3-model merges out of a group of 6 models fully trained from different initializations are shown.}
% \vspace{-5pt}
\label{fig:mismatches_permute}
\end{figure}

In Fig. \ref{fig:mismatches_permute} we present the percent of neurons getting mismatched in this way by Permute for all merging layers inside a ResNet20 trained on CIFAR100. These results help explain why permutations-based methods suffer a drastic drop in accuracy as the number of models being merged increases since there are severe mismatches between the features being aligned that get compounded.

\subsection{The direct and indirect matching matrices ($\mathbf{T}_\mathcal{CB}$ and $\mathbf{T}_\mathcal{CAB}$ resp.) are closer for CCA Merge than for Permute}
\begin{figure}[h]
\begin{center}
\includegraphics[width=0.75\textwidth]{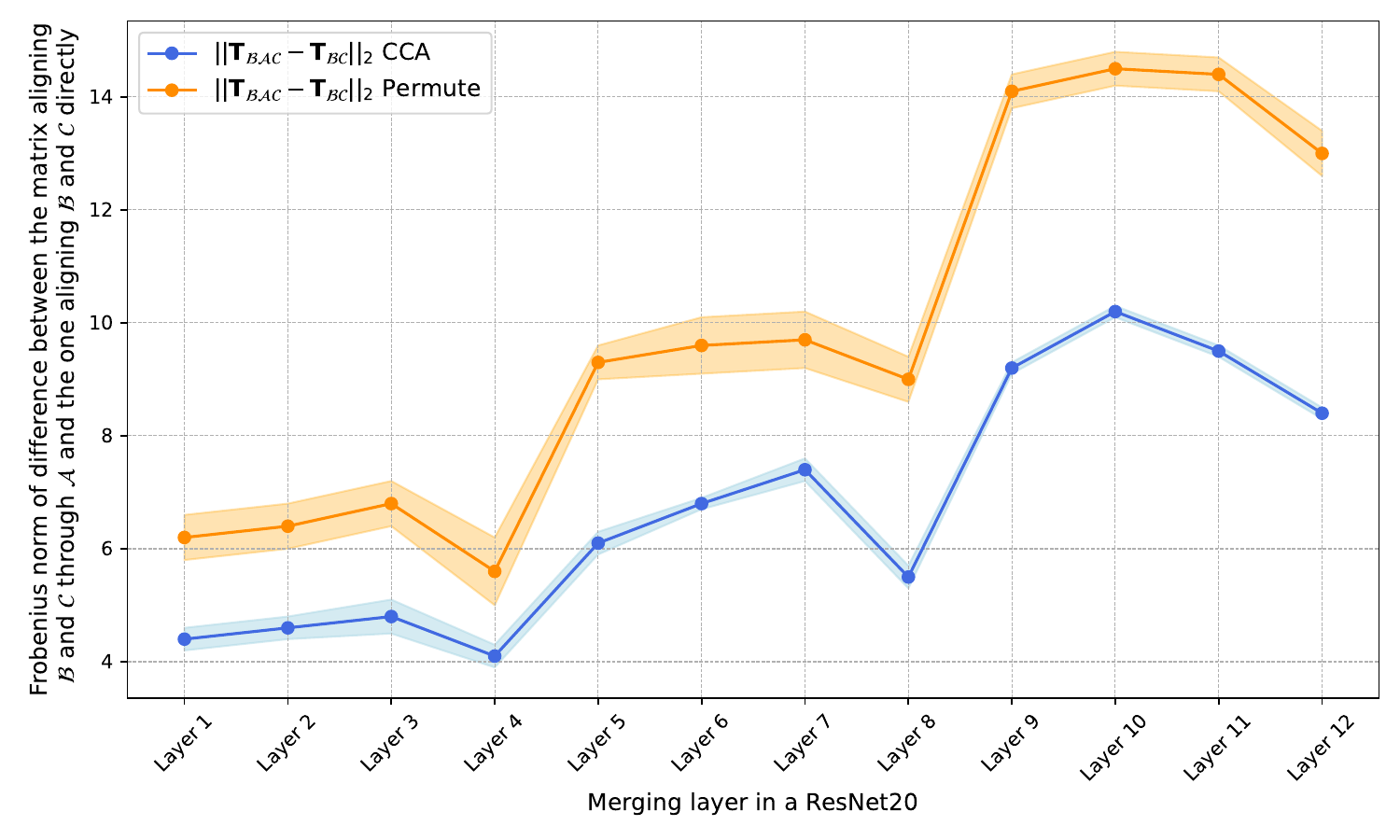}
\end{center}
% \vspace{-12pt}
\caption{Frobenius norm of the difference between the transformation matrices $\mathbf{T}_{\mathcal{B}\mathcal{A}\mathcal{C}}$, aligning $\mathcal{B}$ and $\mathcal{C}$ through $\mathcal{A}$, and $\mathbf{T}_{\mathcal{B}\mathcal{C}}$ aligning $\mathcal{B}$ and $\mathcal{C}$ directly. The mean and standard deviation across 20 possible 3-model merges out of a group of 6 models fully trained from different initializations are shown.}
% \vspace{-5pt}
\label{fig:frob_diffs}
\end{figure}

We can also look directly at the Frobenius norm of the difference between $\mathbf{T}_\mathcal{CAB}$ and $\mathbf{T}_\mathcal{CB}$ for CCA Merge and Permute to see for which method the indirect matching between $\mathcal{C}$ and $\mathcal{B}$ ($\mathbf{T}_\mathcal{CAB}$) and the direct matching between $\mathcal{C}$ and $\mathcal{B}$ ($\mathbf{T}_\mathcal{CB}$) are the most similar. We report the results for the same 20 merges considered above in Fig. \ref{fig:frob_diffs}. We see that the indirect and direct matching matrices between $\mathcal{C}$ and $\mathcal{B}$ are significantly closer for CCA Merge than for Permute, which helps explain why CCA Merge outperforms permutation-based methods in the multi-model setting.

We have run these analyses using the Permute matching method since it is the most straightforward to analyze and we have seen from our empirical results that it constitutes a strong baseline. However we expect all permutation-based methods to be susceptible to these types of non-optimal merging and mismatches because of their permutation matching matrices and since their accuracies behave similarly as the number of models being merged increases (for the multi-model scenario).

\section{Analysis of computational costs}\label{a:comp_costs}
We have tracked the runtime for 5 different 2-model merges of ResNet20x8 models trained on CIFAR100 and we report these values in the table below.

\begin{table}[!ht]
\caption{Runtime for 5 different 2-model merges of ResNet20x8 models trained on CIFAR100}
\vspace{-10pt}
\label{t:compute_costs}
\begin{center}
\begin{tabular}{l|l l l l}
            & Permute & ZipIt! & CCA Merge & Matching Weights\\
            \hline
            Time for the whole merging (s)	& 33.63±0.08 & 37.39±0.34	& 34.50±0.57 & 3.07±0.27\\
Time for computing the transforms (s) & 0.05±0.01 & 3.77±0.35 & 0.93±0.57\\
        \end{tabular}
\end{center}
\end{table}

The merging for methods relying on data (Permute, ZipIt!, CCA Merge) can be split into 2 parts, computing the metrics (eg. covariances or correlations) and computing the transformations. Computing the metrics is by a large margin the most time-expensive part of the procedure, taking on average 33.44s when using the entire CIFAR100 training set. Computing the transforms on the other hand takes less than a second for Permute and CCA Merge and 3.77s for ZipIt!, which represents $\leq3\%$ (for Permute and CCA Merge) and ~10\% (for ZipIt!) of the time required for the entire merging. Among all data-based merging methods, CCA Merge performs the best in accuracy with comparable computation time, therefore it should be prioritized over other such methods. Matching Weights, which doesn't require data, takes ~3.07s to complete, however it performs worse than CCA Merge in terms of accuracy in practice.

It is also valuable to describe how these costs scale with model and dataset size. Computing the correlations scales quadratically with the number of neurons in a layer and linearly with the dimension of the activations (which takes into account the size of the input images and the number of examples used to compute the metrics).

\section{Merged Models vs. Endpoint Models Accuracies}\label{a:merged_vs_endpoints}
In past works such as \citet{ainsworth2023_git-rebasin} or \citet{jordan2023repair}, the merged ResNet models achieving the same accuracy as the endpoint models (or close to) when training on the full train datasets have been extremely wide ones. Specifically, for the ResNet20 architecture on CIFAR10 those results were obtained for models of widths $\times16$ or greater. Zero barrier merging was not achieved for the model widths considered in our work. Exploring the very wide model setting is not an objective of ours since the effect of model width on merging is already well understood, with wider models leading to better performing merges \cite{entezari2022_perm-invariance-lmc, ainsworth2023_git-rebasin, jordan2023repair}. Therefore we only trained models of width up to $\times8$, which are more likely to be encountered in practice. For the model widths reported in our work the accuracy barriers are consistent with those reported in \citet{jordan2023repair} for Permute (solving the linear sum assignment problem maximizing the correlation between matched neurons), with CCA Merge outperforming those results.

Furthermore, our ResNet20 results are reported on the CIFAR100 dataset which is a harder classification task than CIFAR10, therefore it is harder to achieve zero-barrier merging. Also, in all the disjoint training scenarios presented in Table \ref{t:split_data_results} we do achieve greater accuracy than the endpoint models and outperform past methods, not only in settings considered by past works such as 80\%-20\% training splits but in novel settings as well, such as Dirichlet training splits.
\end{document}